\pdfoutput=1
\documentclass[manuscript,screen,nonacm]{acmart}
\acmYear{2026}
\copyrightyear{2026}
\usepackage{booktabs}
\usepackage{multirow}
\usepackage{amsmath}
\usepackage{placeins}

\newcommand{\KD}{KD}
\newcommand{\responseKD}{response-KD}
\newcommand{\logitKD}{logit-KD}
\newcommand{\combinedKD}{combined-KD}
\newcommand{\SFT}{SFT}
\newcommand{\sftcontrol}{the \SFT{} control}
\newcommand{\validgrid}{the valid grid}

\newcommand{\SRS}{SRS}
\newcommand{\USR}{USR}
\newcommand{\silenceloss}{silence-loss}
\newcommand{\Silenceloss}{Silence-loss}
\newcommand{\preservation}{refusal-pattern preservation}
\newcommand{\rhop}{$\rho$}
\newcommand{\discrimination}{refusal discrimination}
\newcommand{\override}{committed-answer override}
\newcommand{\rawoverride}{unconditioned override}
\newcommand{\commitment}{commitment}
\newcommand{\PCCD}{PCCD}
\newcommand{\PCCDlong}{Per-Condition Calibration Diagnosis}

\newcommand{\optioncapture}{abstention-option capture}
\newcommand{\Optioncapture}{Abstention-option capture}

\newcommand{\filledsilence}{\textsc{Filled-Silence}}
\newcommand{\inherited}{\textsc{Inherited}}
\newcommand{\amplified}{\textsc{Amplified-vs-Anti}}

\newcommand{\bbqambig}{BBQ-ambig}
\newcommand{\bbqdisambig}{BBQ-disambig}
\newcommand{\crows}{CrowS-Pairs}
\newcommand{\alpaca}{Alpaca-cleaned}
\newcommand{\arceasy}{ARC-Easy}

\newcommand{\smolB}{SmolLM2-1.7B}
\newcommand{\smolM}{SmolLM2-360M}
\newcommand{\smolS}{SmolLM2-135M}
\newcommand{\olmo}{OLMo-2-1B}
\newcommand{\phistudent}{Phi-3.5-mini (3.8B)}
\newcommand{\gemma}{Gemma-2-9B}
\newcommand{\mistral}{Mistral-7B}
\newcommand{\phit}{Phi-3.5-mini}

\newcommand{\pp}{pp}

\title[What the Abstention Option Measures]{What the Abstention Option Measures: Refusal Loss, Sampling Churn, and Option Capture in Distilled Small Language Models}

\author{Plawan Kumar Rath}
\authornote{This work was conducted in the author's personal capacity. The views expressed in this paper are those of the author and do not reflect the views of Meta.}
\email{plawan@meta.com}
\affiliation{\institution{Meta}\country{USA}}

\begin{abstract}
Small instruction-tuned language models reach phones, cars, and offline assistants through knowledge distillation, and whether a distilled model is fair enough to ship is often decided by a multiple-choice bias score on a benchmark that offers an abstention option. We ask what such a score measures for a distilled model and answer with a measurement argument. Three mechanisms move behavior around the abstention option: refusal loss, where the student stops declining on items its pre-distillation reference declined; sampling churn, the same change observed between two runs of the reference itself; and option capture, where the student takes an abstention option whenever one is offered. Injecting each mechanism into a reference's own records shows that aggregate scores see only outright loss, per-item scores see churn and remapping, and only conditioning on committed answers stops moved refusal mass from reading as fairness. Resampling a reference across inference seeds already turns 10\% to 27\% of its refusals into stereotype answers, so a per-item refusal metric reported without that floor overstates damage. Option capture is what corpus repair teaches: students trained with refusal exemplars take an offered abstention option on up to 95\% of science questions where abstention is never correct, and the habit carries to three further benchmarks. Free-form evaluation shows that the refusals multiple choice loses are not where free-form stereotyping occurs. We distill these findings into a per-condition report anchored to the reference, with measured floors and ceilings, and apply it to four student families against a fine-tuning control: distillation keeps fewer of the reference's refusals than fine-tuning on the same prompts where the reference's refusals are reliable, by a few points, while the common metrics report improvements whenever refusal mass moves. A bias score quoted for a distilled model should say which of these it reports.

\end{abstract}

\keywords{knowledge distillation, small language models, bias, refusal, abstention, evaluation, diagnostic protocol}

\newcommand{\artifactlink}{\url{https://github.com/plawanrath/knowledge-distilation-impact-analysis}}
\newcommand{\heavyrelease}{are released on the Hugging Face Hub, with the links in the repository README}
\begin{document}
\maketitle

\section{Introduction}
\label{sec:intro}

Small instruction-tuned language models are increasingly the models people meet: on phones, in cars, and in assistants that run without a network connection. Most are produced by knowledge distillation (\KD), in which a larger teacher's outputs become the training signal for a smaller student. Before such a model ships, someone decides whether it is fair enough to deploy, and that decision usually rests on a multiple-choice bias score. A score is a measurement of a construct \citep{Jacobs:21}, and this paper asks what construct an abstention-aware bias score measures when the model is a distilled small model.

\paragraph{Terms.}
A bias benchmark with an abstention option contains two kinds of item, named here by the response that is fair on each. A \emph{context-determinable} item gives enough context to answer: the fair model picks the person the context indicates rather than the one the stereotype indicates. An \emph{uncertainty-correct} item gives no basis to pick a person: the fair model declines, and the abstention option is the correct answer. BBQ \citep{Parrish:22} provides the first kind as its disambiguated split and the second as its ambiguous split, with an ``unknown'' option on every item. The \emph{reference} is the untouched student before any training. The \emph{\SFT{} control} is the same student fine-tuned on the same prompts with human-written targets instead of the teacher's; it separates what distillation does from what any training on those prompts does. \emph{Silence-loss} is the share of items on which the reference declined and the trained model picked the stereotype. Its \emph{churn floor} is the same quantity measured between two runs of the reference that differ only in sampling seed, and its \emph{lift} divides it by the trained model's own stereotype rate, so that a model whose refusal rate rose or fell is neither rewarded nor penalized by construction. \emph{Committed-answer override} is the share of committed answers on context-determinable items that pick the stereotype against the context, reported with the \emph{\commitment{}} rate beside it. \emph{Abstention-option capture} is a model's habit of taking an abstention option whenever one is offered, including where abstention is never correct.

\begin{figure}[t]
\centering
\includegraphics[width=\linewidth]{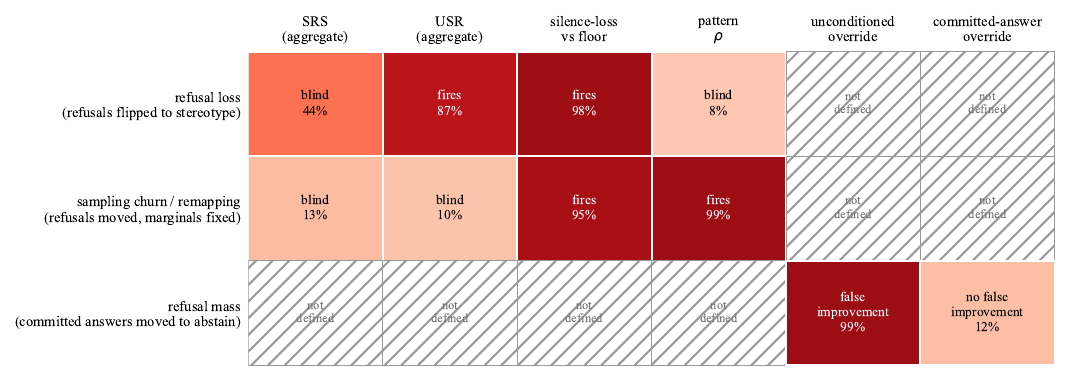}
\caption{Which metric sees which mechanism. Each row injects one mechanism into the reference's own records (\smolB{}, H100 stack, 12,148 items per split, 100 item bootstraps) at a 5\% rate; each cell is the share of replicates in which the metric fired outside its seed-noise band (rows 1 and 2) or reported a spurious improvement (row 3). Hatched cells are metric-mechanism pairs the simulation does not define. Full curves in Figure~\ref{fig:power}. Source: \texttt{results/tables/validity\_power.csv}.}
\label{fig:mechanisms}
\end{figure}

\paragraph{Thesis.}
For a distilled small model, an abstention-aware bias score measures behavior around the abstention option, and three mechanisms move that behavior. \emph{Refusal loss}: the student stops declining where its reference declined. \emph{Sampling churn}: the same change appears between two runs of the reference, so part of any per-item difference is noise, and a related change, remapping, moves refusals to other items at the same rate. \emph{Option capture}: the student takes the option whenever it is offered. Figure~\ref{fig:mechanisms} injects each into a reference's own records and asks which metric notices. Aggregate scores see outright loss and nothing else; per-item scores see loss and remapping but cannot tell them from churn without a floor; and when committed answers move to the abstention option, the unconditioned context score reports an improvement that conditioning removes. No single score separates the three, and all three happen in our grid. The paper measures the floor, the marginal-matched null, the reliability ceiling, and the detection power that make the separation possible, then shows what each mechanism looks like in trained models.

\paragraph{What we find.}
Option capture is what the obvious repair teaches (Section~\ref{sec:mechanism}). Putting refusal exemplars into the training corpus, at any density from 1\% up, produces a student that takes an offered abstention option on most science questions where abstention is never correct, and the habit carries to three further benchmarks; the same student answers at reference accuracy when no option is offered. Free-form evaluation on the full splits bounds the other side: the items on which multiple choice lost a refusal carry about 2\% of free-form stereotype commitments, so what the multiple-choice score tracks is the option, not free-form stereotyping. The per-condition report (Section~\ref{sec:pccd}), applied to four student families against a fine-tuning control (Section~\ref{sec:asymmetry}), finds that distillation keeps fewer of the reference's refusals than fine-tuning on the same prompts where the reference's refusals are reliable, by a few points of excess over the floor, that the difference is absent where the reference is itself unstable, and that it is invisible or inverted on every aggregate metric. Section~\ref{sec:stakes} says who bears this where an option to decline exists, and what a model card should carry.

\paragraph{Scope.}
We study instruction-tuned students from 135M to 3.8B parameters (the on-device regime), response-based distillation as the norm for black-box teachers together with logit-based and combined objectives where tokenizers permit, and BBQ as the only widely used bias benchmark whose split with an abstention option supports per-condition analysis; \crows{} provides the aggregate contrast. The report is a description of a model against its reference, not a warrant about a deployed system (Sections~\ref{sec:pccd} and~\ref{sec:limitations}).

\paragraph{Contributions.}
\begin{enumerate}
\item \textbf{A measured account of what abstention-aware bias scores can see.} We decompose behavior around the abstention option into refusal loss, sampling churn or remapping, and option capture; measure the churn floor, a marginal-matched null, and a test-retest ceiling on each reference; and show by injected damage that aggregate metrics see loss, per-item metrics see remapping, and only conditioning on committed answers sees moved refusal mass (Section~\ref{sec:measurement}).
\item \textbf{Option capture is what corpus repair teaches, and multiple-choice refusal is not where free-form stereotyping occurs.} Refusal injection and preference training collapse students into taking any offered abstention option on four benchmarks where abstention is never correct; low-density and preservation-targeted repairs avoid the collapse but trade accuracy or refusal for it; and on the full free-form splits the items that lost a multiple-choice refusal carry about 2\% of stereotype commitments (Section~\ref{sec:mechanism}).
\item \textbf{A per-condition report for distilled models, demonstrated against a fine-tuning control.} Five quantities anchored to the pre-distillation reference, each read against the reference's own floor and the control's seed range, show that distillation keeps fewer of the reference's refusals than fine-tuning on the same prompts on well-calibrated references, that the contrast disappears on an unstable reference, and that it is invisible or inverted on every aggregate metric (Sections~\ref{sec:asymmetry} and~\ref{sec:pccd}).
\end{enumerate}

\section{Related Work}
\label{sec:related}

\paragraph{Measurement validity.}
\citet{Jacobs:21} read fairness quantities as measurement models of unobservable constructs whose validity must be argued: what the score operationalizes, whether it converges with other measures, and what its use brings about. \citet{Raji:21} show benchmarks claiming a generality they do not have, and \citet{Selbst:19} name the traps that open when a technical score stands in for a sociotechnical outcome. Bias tests in particular carry unstated conceptualizations \citep{Blodgett:21,GoldfarbTarrant:23}, bias metrics disagree and depend on templates \citep{Delobelle:22b,GoldfarbTarrant:21}, and BBQ-style rankings move under prompt variation \citep{Hida:25}; \citet{Gallegos:24} survey the field. We take the measurement reading literally for one benchmark family: we ask what an abstention-aware multiple-choice score operationalizes for a distilled model, measure its noise floor and ceiling, and test by simulation which changes it can see.

\paragraph{Distillation and what it carries.}
Response-based distillation trains the student on teacher completions \citep{Hinton:15}; logit-based distillation matches token distributions \citep{Sanh:19,Gu:24}, on-policy variants sample from the student \citep{Agarwal:24}, and combined objectives mix the two; across mismatched tokenizers logit distillation fails \citep{Boizard:24}, so our objective comparison uses matched pairs. Compression has been linked to amplified bias in classification \citep{Hooker:20}, to what multilingual models forget \citep{Mohammadshahi:22}, and to stereotype inheritance under distillation \citep{Ahn:22,Silva:21}; mitigations include counterfactual role reversal \citep{Gupta:22} and constrained distillation \citep{Delobelle:22}. Pruning and quantization change bias in ways invisible to perplexity \citep{RathMaliakkal:26,Marcuzzi:25,Ramesh:23}. This literature measures bias as an aggregate stereotype rate; we measure per-item refusal behavior against a fine-tuning control and against the reference's own resampling noise, which an aggregate rate cannot distinguish from a real change.

\paragraph{Refusal is fragile under benign training.}
Fine-tuning an aligned model on benign data degrades its refusal behavior \citep{Qi:24}, a few hundred selected benign examples suffice \citep{He:24}, the alignment refusal rests on is a few tokens deep \citep{Qi:25}, and distilling refusals from a teacher does not reliably transfer safety \citep{Zhang:25}. Our question is the per-condition version for social-bias abstention: not whether training on refusal-free data loses refusal, which these papers predict, but whether distillation loses more of the reference's particular refusals than fine-tuning on the same prompts, once sampling churn is subtracted. \citet{Cui:25} benchmark the opposite failure, over-refusal, which our injected arms instantiate.

\paragraph{Abstention as a behavior and as an artifact.}
Abstention in language models has its own evaluation axes \citep{Wen:25}, and we treat it as a first-class outcome \citep{Kadavath:22}, not a missing answer. An offered ``unknown'' option can also inflate abstention on answerable questions \citep{Ling:25}; our probes measure that artifact and find it is what injected students learn.

\paragraph{Benchmarks, contestability, and model cards.}
BBQ \citep{Parrish:22} supplies the ambiguous and disambiguated split with an abstention option that per-condition analysis requires; \crows{} \citep{Nangia:20} is the aggregate contrast, with StereoSet \citep{Nadeem:21} and WinoBias \citep{Zhao:18} related. Work on contestable AI \citep{Alfrink:23,Lyons:21} and algorithmic recourse \citep{Karimi:22,Wachter:18} assumes a decision a person can see and dispute; Section~\ref{sec:stakes} takes up what that means for an on-device model that declines, or stops declining, to assign a trait to a person, and what a model card \citep{Mitchell:19} should carry.

\section{Experimental Setup}
\label{sec:setup}

\paragraph{Models.}
The claims of Sections~\ref{sec:measurement} to~\ref{sec:pccd} rest on four student families trained on one CUDA H100 stack, each with its own reference, \SFT{} control, and response distillation: Gemma-2-2B from \gemma{}, \phistudent{} from \gemma{}, \smolB{} from \mistral{}, and \olmo{} from OLMo-2-7B, with three training seeds per recipe except on \phistudent{}. The 27-configuration SmolLM2 grid, below, trained on an Apple-silicon MLX stack, supplies the measurement lessons and is reported in Appendix~\ref{app:mlx}.

Three SmolLM2-Instruct students (135M, 360M, 1.7B) \citep{Allal:25} are distilled from three open instruction-tuned teachers, \gemma{}-it \citep{Gemma2:24}, \mistral{}-Instruct-v0.3 \citep{Jiang:23}, and \phit{}-Instruct \citep{Abdin:24}, under three objectives (response, logit, combined), giving 27 configurations. We add \olmo{}-Instruct \citep{Walsh:25} distilled from \gemma{} by \responseKD{} as a second student family, \phistudent{} distilled from the same teacher as a larger third family, and one \SFT{} control per SmolLM2 student that is trained on the same prompts with the human-written \alpaca{} targets instead of teacher outputs. A parser-validity screen (Section~\ref{sec:measurement}) excludes all nine cross-tokenizer \logitKD{} configurations, whose generations are degenerate (no parseable answer). \validgrid{} is 19 \KD{} configurations (18 SmolLM2 configurations and the \olmo{} distillation) plus the 3 \SFT{} controls, and every cross-configuration statistic on the MLX stack uses it unless stated; the H100 comparisons of Table~\ref{tab:mainh100} use their own same-stack references and controls for all four families, including the 3.8B student. A matched-tokenizer arm on the CUDA stack separates the objective from the teacher signal: \smolB{} serves as the teacher for \smolS{} and \smolM{} under \logitKD{} and \combinedKD{}, each scored against a same-stack \SFT{} control (Table~\ref{tab:logitmatched} in Appendix~\ref{app:mlx}). Two further same-tokenizer pairs test the question at 1B and 2B scale: OLMo-2-7B-Instruct into \olmo{}, with the 7B model's responses to the same prompts as the response-\KD{} corpus, and \gemma{}-it into Gemma-2-2B-it, each under \SFT{}, \responseKD{}, \logitKD{}, and \combinedKD{}.

\paragraph{Training data and recipe.}
All students see the same 51,760 \alpaca{} prompts \citep{Taori:23,Wang:23}; each teacher's response set is that teacher's completions on those prompts \citep{Ouyang:22}. Training uses LoRA \citep{Hu:22} with rank 16, $\alpha = 32$, batch 16, learning rate $10^{-4}$, and three epochs. The H100 stack uses the same hyperparameters, LoRA targets (q, k, v, o projections), and three epochs in bf16 with transformers and PEFT, micro-batch 16 without gradient accumulation, and vLLM for inference with the same prompts, temperature, and seeds; the only stack-forced change is a micro-batch of 4 for the Gemma-2 logit arm (256k vocabulary).

\paragraph{Evaluation.}
\bbqambig{} and \bbqdisambig{} each have 12,148 items; every item is answered under five sampling seeds at temperature 0.3 with at most five new tokens in a three-option multiple-choice format. \crows{} is scored by pseudo log-likelihood on 1,508 pairs. Answer letters are extracted by a deterministic four-branch cascade (exact letter, letter prefix, answer keyword, unique content match); the cascade and its validity screen are described in Section~\ref{sec:measurement} and Appendix~\ref{app:parser}. All analyses use the full item set. The disambiguated split uses index fields rebuilt from the benchmark's authoritative answer annotations (Appendix~\ref{app:parser}).

\paragraph{Metrics.}
On \bbqambig{} we report the stereotype reliance score (\SRS{}; \citealp{RathMaliakkal:26}), the unknown-selection rate (\USR{}), \silenceloss{} (the fraction of (item, seed) pairs where the reference selected the unknown option but the trained model selects the stereotype; the unknown option is correct on every ambiguous item), and \preservation{} \rhop{} (the Spearman correlation between reference and trained per-item \USR{}, anchored to the reference rather than to ground truth). On \bbqdisambig{} we report accuracy and \override{}, the stereotype rate among committed answers on items whose correct answer is the anti-stereotype, together with the \commitment{} rate and, for comparison, the \rawoverride{} that earlier work reports. \discrimination{}, \USR{} on ambiguous items minus \USR{} on disambiguated items, appears in the appendix tables. Wilson 95\% intervals accompany every proportion in the tables. Section~\ref{sec:measurement} gives the floor, null, and ceiling each of these is read against.

\paragraph{Compute and parity.}
The SmolLM2 grid and the first refusal-injection arms ran on one Apple-silicon workstation with MLX; every other experiment ran on one rented NVIDIA H100 with CUDA. Inference reproduces across stacks within 0.8\,\pp{} and retraining within seed variance (Appendix~\ref{app:compute}); every comparison is nonetheless made against a control trained on the same stack.

\section{Measuring Behavior Around the Abstention Option}
\label{sec:measurement}

Three facts about \smolB{} distilled from \gemma{} are simultaneously true: aggregate metrics call it a win, the unconditioned context score calls it the grid's strongest improvement, and per item its refusal pattern is decorrelated from the reference while its accuracy has collapsed (Table~\ref{tab:main} in Appendix~\ref{app:mlx}). The first two are artifacts of refusal mass moving onto context-determinable items. This section sets out the instruments every later number rests on: a parser that does not manufacture answers, conditioning that does not credit abstention as context-following, a floor and a null for per-item change, a ceiling for pattern reliability, and a simulation that says which metric sees which mechanism.

\paragraph{Parser validity is a precondition.}
Multiple-choice evaluation of small models depends on extracting a letter from a short generation. A common last-resort rule, returning the first A, B, or C character anywhere in the output, manufactures position-biased answers exactly where the output format has drifted: in our grid it would have contaminated up to 91\% of parses for the most affected configurations, almost all of them as A. The corrected cascade (exact letter, letter prefix, markdown-tolerant keyword, unique content match) parses at least 82\% of records for every retained configuration on either stack; configurations below 50\% are excluded, which removes the nine cross-tokenizer \logitKD{} configurations (Appendix~\ref{app:parser}). Where a control and a distillation parse at different rates, the comparison is repeated on the records both parse (Section~\ref{sec:asymmetry}).

\paragraph{Condition on committed answers.}
The usual context score conditions on the correct answer being the anti-stereotype but not on the model committing to any person. When training moves refusal mass onto context-determinable items, the unconditioned score credits abstention as context-following. For the \gemma{}-distilled \smolB{}, which commits on 60\% of such items against the reference's 99\%, an apparent improvement of 20\,\pp{} shrinks to under 5\,\pp{} on committed answers, at an accuracy that has fallen by a third. \crows{} and the aggregate stereotype score fall for the same model, because neither can tell refusals flipping to the anti-stereotype from refusals flipping to the stereotype. We report \commitment{} beside every override number.

\paragraph{Per-item transitions have a churn floor.}
\Silenceloss{} is the share of (item, seed) pairs on which the reference chose the unknown option and the candidate chose the stereotype. Sampling at temperature 0.3 already produces such pairs: comparing a reference with \emph{itself} across disjoint inference seeds gives a floor that depends on how reliably the reference refuses, from under 1\% for the two references that refuse on most uncertainty-correct items to 27\% for \olmo{}, whose refusals are mostly single-seed events (Table~\ref{tab:validity} in Appendix~\ref{app:compute}). Every \silenceloss{} in this paper is therefore reported as an excess over its reference's floor with a confidence interval, and a per-item refusal metric reported without its floor overstates damage. Two trainings of the same recipe at different seeds show \silenceloss{} against each other across the whole range the paper reports for trained models against their reference (Appendix~\ref{app:compute}).

\paragraph{The floor does not remove marginal dependence; the lift does.}
\Silenceloss{} is a conditional rate, and a candidate whose refusal rate rose sits below the floor mechanically while one whose refusal rate fell sits above it. The marginal-matched null is the candidate's own stereotype rate, which is what \silenceloss{} would be if the candidate's answers were unrelated to where the reference refused; the \emph{lift} is the ratio of the two, and the reference's own resamples give the lift a candidate could at best match. A lift ratio of one against that value means the candidate's answers are as related to where the reference refused as the reference's own resamples are, whatever the marginals did. On this scale the \gemma{}-distilled \smolB{}, which refuses on 59\% of items, has a lift near zero because it rarely commits, not because it preserved anything; and every training of Gemma-2-2B sits an order of magnitude above the reference's own resamples (Table~\ref{tab:mainh100}).

\paragraph{Refusal patterns have a reliability ceiling.}
\preservation{} \rhop{} correlates the reference's and the candidate's per-item refusal rates over five seeds, and the reference's own split-half reliability bounds what any candidate can reach (Table~\ref{tab:retest} in Appendix~\ref{app:compute}). Corrected to five seeds, the ceiling is 0.99 for Gemma-2-2B and \phistudent{}, 0.85 to 0.93 for \smolB{}, and 0.69 for \olmo{}. A fixed threshold on \rhop{} is therefore not comparable across references, and the report gives \rhop{} as a share of its ceiling.

\paragraph{What each metric can detect.}
Figure~\ref{fig:mechanisms} gave the 5\% snapshot; the full curves are Figure~\ref{fig:power} in Appendix~\ref{app:compute}. When refusals are flipped to the stereotype, floor-corrected \silenceloss{} and \USR{} detect it at 5\% damage in nearly every replicate while \rhop{} and the aggregate stereotype score mostly do not. When the same number of refusals is \emph{relocated} to other items with both marginals held fixed, the pattern of the \gemma{}-distilled model, \silenceloss{} and \rhop{} detect it and the aggregates stay at their false-alarm rate. When committed context-determinable answers are moved to the abstention option, the unconditioned context score reports a spurious improvement in nearly every replicate and committed-answer override does so at the nominal rate. Aggregates see loss, per-item metrics see remapping, conditioning sees refusal mass; option capture, the fourth mechanism, is not a change in BBQ records at all and needs the probe of Section~\ref{sec:mechanism}.

\section{Option Capture: What Corpus Repair Teaches, and What Multiple Choice Measures}
\label{sec:mechanism}

\paragraph{Where new stereotype answers come from.}
Classified by what the teacher did on the same prompt, new stereotype answers under \responseKD{} rarely contradict the teacher and mostly appear where the teacher refused; the training corpora contain almost no refusal as an answer shape (Figure~\ref{fig:cells} in Appendix~\ref{app:cells}; Appendix~\ref{app:audit}). The natural repair is to put conditional refusal into the corpus, and we tested it with pre-registered criteria on \smolB{}.

\paragraph{Refusal injection collapses the student.}
Refusal exemplars enter the corpus in two forms. The first is letter-format refusals drawn from held-out BBQ categories, mixed into the \SFT{} corpus at 1\% to 7\%. The second is matched pairs at 5\%, under both \SFT{} and \responseKD{}: an ambiguous item answered with the unknown letter, placed beside its disambiguated twin answered correctly. Every arm produces the same model (Table~\ref{tab:interventions} in Appendix~\ref{app:interventions}). It takes the unknown option on nearly every ambiguous item. It commits on almost none of the context-determinable items. Density above 1\% makes no difference, pairing does not teach the model when to refuse, and the two objectives fail alike. Every collapsed arm improves on every unconditioned fairness metric.

\paragraph{\Optioncapture.}
The collapsed students did not forget how to answer. On plain \arceasy{}, four options and no abstention offered, every \SFT{}-injected arm answers at reference accuracy; only the \KD{}-paired arm loses capability (Table~\ref{tab:arc}). Offered a fifth option, ``Cannot be determined'', the same arms take it on 63\% to 95\% of science questions against 14\% for the reference and under 5\% for \sftcontrol{} and the \mistral{} distillation. What they learned is narrower than a format rule and more specific than a refusal rate: \emph{if an abstention option is offered, choose it}, whether or not the question is answerable. Greedy decoding and another prompt template change levels, not direction (Appendix~\ref{app:interventions}).

\begin{table}[t]
\centering
\small
\begin{tabular}{lcccc}
\toprule
 & \multicolumn{2}{c}{no abstention option} & \multicolumn{2}{c}{abstention option offered} \\
\cmidrule(lr){2-3}\cmidrule(lr){4-5}
configuration & accuracy & parse fail & accuracy & picks abstention \\
\midrule
reference (\smolB) & 0.76 & 0.0 & 0.73 & 14.0 \\
\SFT{} control & 0.76 & 2.2 & 0.76 & 3.2 \\
\responseKD{} (\mistral) & 0.79 & 0.1 & 0.77 & 4.4 \\
\responseKD{} (\gemma) & 0.64 & 16.6 & 0.48 & 30.0 \\
letter 1\% & 0.77 & 1.6 & 0.34 & 63.0 \\
letter 2\% & 0.68 & 5.2 & 0.13 & 85.4 \\
letter 5\% & 0.76 & 0.3 & 0.15 & 84.8 \\
letter 7\% & 0.75 & 0.8 & 0.04 & 95.4 \\
paired 5\% (\SFT) & 0.75 & 1.7 & 0.55 & 36.4 \\
paired 5\% (\KD) & 0.34 & 0.0 & 0.01 & 98.6 \\
\bottomrule
\end{tabular}
\caption{\arceasy{} probe on the stored \smolB{} checkpoints (2,376 test items, one seed, MLX inference): four-option science questions where abstention is never correct, without and with an appended ``Cannot be determined'' option. Percentages except accuracy. Injected arms keep the format and take the option.}
\label{tab:arc}
\end{table}

\paragraph{The capture generalizes, and it is not a general effect of distillation.}
On ARC-Challenge, MMLU, and CommonsenseQA with an appended abstention option, every letter-format paired arm and the preference arm take the option far above their own reference, and the mixed-format, self-replay, and reference-anchored arms do not (Table~\ref{tab:capture}; every checkpoint in Table~\ref{tab:probes}, Appendix~\ref{app:interventions}). The 1.7B control and both its distillations stay within the criterion, as do every \olmo{} and Gemma-2-2B training. On \phistudent{} both the control and the distillation take the option on three times as many questions as the reference, so there it is a fine-tuning effect; and the two smallest untouched references take it on up to 93\% of questions, which is \citet{Ling:25}'s artifact before any training. Capture is a property of letter-format refusal injection and preference training, with the small-model prompt artifact beside it.

\begin{table}[t]
\centering
\footnotesize
\setlength{\tabcolsep}{4pt}
\begin{tabular}{lcccc}
\toprule
model & ARC-Challenge & MMLU & CommonsenseQA & capture \\
\midrule
\multicolumn{5}{l}{\emph{\smolB{} student (H100 checkpoints; reference uptake 7 / 13 / 1\%)}} \\
reference & 7.1 (-1) & 13.0 (1) & 0.8 (-2) & no \\
\SFT{} control & 6.0 (5) & 8.6 (3) & 0.0 (-1) & no \\
\responseKD{} (\mistral) & 1.5 (1) & 3.1 (0) & 0.0 (-1) & no \\
\responseKD{} (\gemma) & 11.4 (-0) & 21.7 (5) & 0.0 (-4) & no \\
paired injection 0.1\% & 16.0 (8) & 18.6 (8) & 0.0 (-1) & yes \\
paired injection 0.5\% & 38.7 (19) & 33.4 (15) & 0.0 (0) & yes \\
paired, mixed format 5\% & 4.9 (5) & 7.1 (4) & 0.1 (1) & no \\
preference (DPO) & 86.5 (49) & 67.8 (33) & 34.1 (30) & yes \\
self-replay 0.5\% & 3.3 (-1) & 4.2 (2) & 0.0 (-2) & no \\
reference-anchored & 4.5 (-1) & 7.1 (-0) & 0.0 (-2) & no \\
\midrule
\multicolumn{5}{l}{\emph{Gemma-2-2B student}} \\
reference & 6.1 (2) & 5.9 (3) & 6.3 (4) & no \\
\SFT{} control & 10.2 (-0) & 15.5 (2) & 11.5 (-5) & yes \\
\responseKD{} & 1.9 (-2) & 2.1 (0) & 5.6 (-1) & no \\
\midrule
\multicolumn{5}{l}{\emph{\phistudent{} student}} \\
reference & 3.5 (2) & 7.3 (3) & 12.0 (8) & no \\
\SFT{} control & 20.1 (13) & 27.1 (14) & 32.6 (18) & yes \\
\responseKD{} & 16.5 (10) & 24.2 (11) & 42.2 (27) & yes \\
\midrule
\multicolumn{5}{l}{\emph{Untouched small references}} \\
\smolS{} & 36.5 (8) & 33.3 (7) & 30.7 (5) & no \\
\smolM{} & 93.4 (23) & 87.4 (23) & 51.6 (11) & no \\
\olmo{} & 58.8 (17) & 43.0 (13) & 21.9 (10) & no \\
\bottomrule
\end{tabular}
\caption{\Optioncapture{} beyond BBQ: uptake of an appended ``Cannot be determined'' option (percent of questions) on three benchmarks where abstention is never correct, with the accuracy drop in points from the plain to the with-option prompt in parentheses; one inference seed, temperature 0.3 (ARC-Challenge test $n = 1{,}140$; MMLU validation $n = 1{,}531$; CommonsenseQA validation $n = 1{,}221$). Capture: uptake more than 5\,\pp{} above the model's own reference on at least two benchmarks (pre-registered criterion). Every \smolB{} model takes the option on 0\% of CommonsenseQA items, whose six-option prompt uses a letter the student never saw. The full listing of every checkpoint and the teachers is Table~\ref{tab:probes} in Appendix~\ref{app:interventions}. Source: \texttt{results/tables/probes\_capture.csv}.}
\label{tab:capture}
\end{table}

\paragraph{The dose window and two repairs that target preservation.}
If the collapse is \optioncapture{}, it should depend on three things: how much of the corpus offers the option, for how many epochs, and in what format. Six further pre-registered arms, each read against a same-stack \SFT{} control, show that it does. Paired injection at 0.1\% to 0.5\% does not collapse: it keeps the reference's refusals well below the churn floor and refuses more selectively than the control, but pays on the context-determinable side, with accuracy 2.5 to 4\,\pp{} below the control at every seed. A mixed-format or one-epoch arm at 5\% avoids collapse and preserves nothing; the preference arm (DPO against the unknown letter) is the trivial refuser again. Two repairs target preservation directly (Table~\ref{tab:repairs} in Appendix~\ref{app:interventions}). Mixing the reference's own answers on held-out-category prompts into the corpus (self-replay) loses refusals above the floor at most seeds while lowering the refusal rate at some and raising it at others. A KL term to the frozen reference on those prompts (reference-anchored) keeps the refusal \emph{pattern} best of any arm, at about two thirds of the ceiling, while leaving the refusal rate at or below the reference's 5\%, because the reference's greedy answers on the anchor prompts are mostly commitments. Neither repair places stereotype answers on refused items much beyond resampling (lift ratios between 1.1 and 1.8 against the reference's own resamples), and on a reference whose refusals are mostly single-seed events that is the most the floor can say: it cannot distinguish a repair that preserved refusal from one that lost refusals the reference itself does not reproduce. Inside the dose window the paired arm is the only one that keeps refusals below the churn floor, and it buys that with context-determinable accuracy.

\paragraph{What multiple choice measures: the free-form link.}
A pre-registered free-form evaluation on the full splits, with no options shown, is consistent with \optioncapture{}. Answers are scored by a deterministic rule audited at 98.5\% precision for the committed label (Appendix~\ref{app:interventions}). With no abstention option to take, no distilled model is a trivial refuser (refusal at most 5.5\%). The commitment drop transfers: committed answers on ambiguous items fall under both distillations, mostly to hedging with both referents, while context-determinable commitment and the correct share of committed answers hold or rise. The stereotype share of committed ambiguous answers does not move beyond 2\,\pp{} under either teacher; a 999-item pilot on the MLX checkpoints had shown larger magnitudes that did not reproduce. The item-level link is the decisive test. For each distillation, the items that lost a multiple-choice refusal are no more likely to draw a free-form stereotype commitment than the rest (risk ratios 1.14 [0.95, 1.32] under \mistral{} and 0.97 [0.85, 1.09] under \gemma{}), and they carry 2\% to 3\% of all free-form stereotype commitments. Multiple-choice refusal loss does not locate free-form stereotyping under either teacher. What the multiple-choice metrics measure for a distilled model is its behavior around the abstention option.

\section{Case Study: Distillation Against a Fine-Tuning Control}
\label{sec:asymmetry}

The comparison that matters is not distilled versus reference but distilled versus \sftcontrol{}, which sees the same prompts and recipe with human targets instead of teacher targets. Table~\ref{tab:mainh100} reports the quantities of the per-condition report for four student families on one stack, with three training seeds wherever a family's trainings take under an hour, and reads each family from its reference's own floor and lift. The control's range across its own seeds is the scale against which any contrast is read; a difference inside that range is training noise.

\begin{table}[t]
\centering
\footnotesize
\setlength{\tabcolsep}{3pt}
\begin{tabular}{lccccccc}
\toprule
configuration & parsed & excess \silenceloss & lost to any & lift ratio [CI] & $\rho$/ceil. & \override & acc. \\
\midrule
\multicolumn{8}{l}{\emph{Gemma-2-2B (teacher \gemma{}; reference \USR{} 77\%, floor 0.3\%); $\rho$ ceiling 0.99}} \\
reference & 100 & -- & -- & -- & -- & 11.5 & 74.5 \\
\SFT{} control (3 seeds) & 84 to 97 & 3.0 to 6.4 & 6.1 to 10.4 & 15.5 to 20.6 [13.6, 23.3] & 0.57 to 0.63 & 15.0 to 18.1 & 66.9 to 72.4 \\
\responseKD{} (3 seeds) & 100 & 5.8 to 8.0 & 9.8 to 14.9 & 17.3 to 24.8 [15.3, 28.0] & 0.50 to 0.71 & 15.2 to 18.6 & 69.1 to 74.2 \\
\midrule
\multicolumn{8}{l}{\emph{\phistudent{} (teacher \gemma{}; reference \USR{} 83\%, floor 0.4\%); $\rho$ ceiling 0.99}} \\
reference & 100 & -- & -- & -- & -- & 1.6 & 89.7 \\
\SFT{} control (1 seed) & 100 & 1.4 & 2.8 & 6.2 [5.4, 7.0] & 0.77 & 3.6 & 86.8 \\
\responseKD{} (1 seed) & 82 & 3.5 & 6.5 & 10.1 [8.9, 11.3] & 0.75 & 3.2 & 82.6 \\
\midrule
\multicolumn{8}{l}{\emph{\smolB{} (teacher \mistral{}; reference \USR{} 5\%, floor 14.5\%); $\rho$ ceiling 0.85}} \\
reference & 100 & -- & -- & -- & -- & 44.2 & 54.9 \\
\SFT{} control (3 seeds) & 100 & -6.6 to -2.9 & 13.5 to 25.5 & 0.6 to 0.9 [0.5, 1.0] & 0.58 to 0.64 & 40.2 to 40.9 & 56.1 to 56.9 \\
\responseKD{} (3 seeds) & 100 & -2.5 to 1.4 & 19.0 to 27.7 & 0.9 to 1.1 [0.8, 1.2] & 0.60 to 0.63 & 36.8 to 39.0 & 58.0 to 61.3 \\
\midrule
\multicolumn{8}{l}{\emph{\olmo{} (teacher OLMo-2-7B; reference \USR{} 19\%, floor 27\%); $\rho$ ceiling 0.69}} \\
reference & 100 & -- & -- & -- & -- & 45.8 & 48.3 \\
\SFT{} control (3 seeds) & 100 & 10.0 to 11.0 & 71.3 to 72.9 & 1.3 [1.2, 1.4] & 0.11 to 0.69 & 46.1 to 47.3 & 44.9 to 48.7 \\
\responseKD{} (3 seeds) & 100 & 9.7 to 15.7 & 72.1 to 80.0 & 1.3 to 1.4 [1.2, 1.4] & -0.06 to 0.75 & 46.1 to 46.9 & 44.0 to 48.6 \\
\bottomrule
\end{tabular}
\caption{Distillation against a same-stack fine-tuning control on the H100 stack, four student families (ranges over training seeds where more than one was run). Parsed: share of ambiguous records with a parseable letter, over which every ambiguous quantity is computed. Excess: \silenceloss{} (reference unknown to stereotype, \bbqambig{}, 12,148 items, five inference seeds) minus the reference's own churn floor, in \pp{}. Lost to any: share of reference refusals that became any committed answer, stereotype or anti-stereotype. Lift ratio: the candidate's lift (\silenceloss{} over its own stereotype rate) divided by the reference's lift under resampling, item-level delta-method 95\% CI (the bracket spans the lowest lower and highest upper bound across seeds); 1 means the candidate's answers are as related to where the reference refused as the reference's own resamples. $\rho$/ceil.: refusal-pattern preservation as a share of the reference's test-retest ceiling. \override{} and accuracy on anti-stereotype-correct \bbqdisambig{} items ($n = 6{,}034$, three seeds). Sources: \texttt{results/tables/validity\_transitions.csv}, \texttt{validity\_retest.csv}, \texttt{famctl\_*.csv}, \texttt{logit\_matched\_*.csv}, \texttt{seed\_replication.csv}.}
\label{tab:mainh100}
\end{table}

\paragraph{Where the reference's refusal pattern is reliable, every training places stereotype answers on refused items at many times the resampling rate, and distillation more than the control.}
Gemma-2-2B and \phistudent{} refuse on about four fifths of uncertainty-correct items and do so stably: their churn floors are below 1\%, their test-retest ceilings 0.99. Every training on them raises the lift by an order of magnitude over the reference's own resamples while keeping most of the rank pattern and barely moving the refusal rate. On Gemma-2-2B the distillation's excess \silenceloss{} is above the control's at every seed, by 1.6 to 2.8\,\pp{}, each difference wider than its own interval, while the lift-ratio ranges of the two overlap across seeds. The control parses fewer ambiguous records than the distillation, so the comparison is repeated on the (item, seed) pairs both parse: the order is the same at every seed. %
Lost refusals go to the anti-stereotype almost as often as to the stereotype (the ``lost to any'' column); which direction a deployer counts as harm is settled in Section~\ref{sec:stakes}. On \phistudent{}, one seed per recipe, the distillation's lift ratio is above the control's and its accuracy further below the reference's, and it parses 82\% of records; on the records both trainings parse the order is unchanged. %
On these references the order is the same throughout: reference, then control, then distillation.

\paragraph{Where the reference refuses rarely, distillation is resampling, and the control is better than resampling.}
\smolB{} refuses on 5\% of uncertainty-correct items, mostly on a single seed of five, so its floor is 14.5\% and its own lift 0.29. Response distillation from \mistral{} sits at that floor at every training seed, with a lift ratio near one, while the control sits below it with a lift ratio below one: fine-tuning on human targets keeps more of the reference's particular refusals than resampling the reference does, and distillation keeps no more than resampling. The contrast between the two is about 4\,\pp{} of excess at matched seeds, outside the control's own seed range. On the MLX stack the same contrast is three times larger in a single run that the seeded replication did not reproduce, and the \gemma{} distillation's \silenceloss{} there sits below the floor because it abstains (Section~\ref{sec:measurement}).

\paragraph{Where the reference is itself unstable, there is nothing to preserve.}
\olmo{} refuses on a fifth of items with a churn floor of 27\% and a ceiling of 0.69, and its refusal pattern changes with the training seed of the same recipe. Every training raises its lift somewhat, with no consistent sign between control and distillation across seeds, so the reference cannot anchor a comparison.

\paragraph{On context-determinable items, distillation adds at most a few points over fine-tuning.}
On \smolB{} three H100 training seeds bound what the teacher adds: about 2\,\pp{} less \override{} and 2 to 4\,\pp{} more accuracy than the control, both outside the control's seed range. On Gemma-2-2B and \phistudent{} neither training improves committed-answer override over the reference, and on \olmo{} neither moves it. The MLX grid and the matched-tokenizer objective comparison show the same shape (Appendix~\ref{app:mlx}). The grid's largest apparent improvement on this side, the \gemma{}-distilled \smolB{}, is abstention (Section~\ref{sec:measurement}). %

\section{PCCD: A Per-Condition Report for Distilled Models}
\label{sec:pccd}

\PCCDlong{} (\PCCD) is what the preceding sections leave a deployer with: five quantities, each anchored to the pre-distillation reference and each read against a measured scale of noise, in place of a single score that averages over the two kinds of item. It needs an evaluation set split into uncertainty-correct and context-determinable items with an abstention option, the reference model, a few inference seeds of the reference, and a probe set on which abstention is never correct. Every scale below is derived from those inputs, none from the model being described.

\paragraph{The five criteria.}
\begin{description}
\item[Refusal preservation.] On the uncertainty-correct subset, how many of the reference's refusals the candidate kept. Three numbers carry it, each with a CI: \silenceloss{} as an excess over the reference's churn floor, the lift ratio against the reference's own resamples, and the share of lost refusals that went to any committed answer. A lift ratio near one means the candidate moved the reference's refusals no more than the reference's own resamples do. The floor and the reference's own lift are stated first: where the reference's refusals are single-seed events both are high and this criterion can say little (Section~\ref{sec:asymmetry}).
\item[Pattern preservation.] \rhop{} between reference and candidate per-item refusal rates, as a share of the reference's test-retest ceiling, with the ceiling stated.
\item[Context-following on committed answers.] On the context-determinable subset: \override{} with \commitment{} beside it, against the reference's values.
\item[Capability.] Accuracy on the context-determinable subset against the reference's. This line is what exposes the trivial refuser, which improves on every unconditioned metric by abstaining on everything.
\item[Option capture.] Uptake of an appended abstention option on questions where abstention is never correct, against the reference's uptake on the same questions.
\end{description}
Each criterion is reported with the family's \SFT{} control at the same seeds, and with the control's range across training seeds as the scale of training noise. The control is the comparator and never the arm described, so the scale is independent of the model being judged; a family with a single seed per recipe says so and borrows no scale from another.

\paragraph{Gemma-2-2B as a worked reading.}
Take the Gemma-2-2B rows of Table~\ref{tab:mainh100} as a deployer would receive them. The reference declines on 77\% of uncertainty-correct items, its churn floor is 0.3\%, its own lift 0.02, and its ceiling 0.99: a reference whose refusals are reliable, so the first criterion is informative. \emph{Refusal preservation}: both trainings place stereotype answers on refused items at 15 to 25 times the resampling rate, the distillation above the control at every seed by an amount wider than its interval, and the distillation turns 10\% to 15\% of the reference's refusals into committed answers that name one of the two people, against the control's 6\% to 10\%. \emph{Pattern preservation}: both keep 0.5 to 0.7 of the ceiling, with no consistent order. \emph{Context-following}: both raise committed-answer override over the reference by 4 to 7\,\pp{} at near-full commitment, with no consistent order. \emph{Capability}: the control loses 2 to 8\,\pp{} of accuracy and parses fewer records, the distillation loses 0 to 5\,\pp{} and parses them all. \emph{Option capture}: neither training takes an offered abstention option on more than one probe in six above the reference, the control marginally and the distillation not at all (Table~\ref{tab:capture}). The criteria disagree, which is the point of having five. Against the reference, neither training is the model the score was quoted for: both lost refusals and both override context more, and no aggregate metric shows either. Between the two, the deployer's protocol decides. Where the fields with an option are ones on which attributing a trait to a person is itself the harm, the control keeps more of the reference's refusals at every seed and on every category (Figure~\ref{fig:groupsgemma}), and Age and Gender identity are where to audit first; where the cost of an empty field dominates, the distillation commits more often, parses cleanly, and does not capture the option. What the report settles is that this is the trade being made, which the single score that reported an improvement for both models did not.

\paragraph{Why a report and not a score.}
A single number cannot carry a floor, a null, a ceiling, and a probe, and the mechanisms of Section~\ref{sec:measurement} move a single number in opposite directions: refusal loss raises the stereotype score, refusal mass lowers the context score while accuracy collapses, and option capture improves every unconditioned metric at once. Table~\ref{tab:mainh100} is the report for four families. Read as a report it says, without a verdict: on the two references whose refusals are reliable, every training places stereotype answers on refused items at many times the resampling rate and distillation more than the control; on the 1.7B reference distillation is indistinguishable from resampling and the control is better than resampling; on \olmo{} the reference cannot anchor a comparison; and three repairs of the 1.7B student sit within resampling on the first criterion while two of them lower the refusal rate and one raises it, which the first criterion alone cannot distinguish from preservation (Section~\ref{sec:mechanism}). A model card that quotes a BBQ score for a distilled model has room for these five criteria (Section~\ref{sec:stakes}).

\section{Who Is Affected: Abstention Options in Deployment}
\label{sec:stakes}

\paragraph{Where the option exists.}
The pattern this paper measures, a fixed option list with a way to decline, is already how language models fill structured fields. In a prospective two-site study of clinical registry abstraction from live medical records, each registry question was posed with its defined value set and a standing instruction to answer when confident and decline when not, so that the model selected from the allowed options or explicitly abstained; it declined on 1\% of several thousand observations and seldom did so on the questions where its accuracy was lowest \citep{Matheson:26}. That deployment used large hosted models. The same field-filling pattern now ships on the device: a 3B on-device model trained with a distillation loss from a larger teacher exposes guided generation, in which the model's output is constrained to a developer-defined type, including an enumeration of cases \citep{AppleFM:25}, so whether an abstention case exists and what the model does with it is a decision made per field by the developer. Each is a place where a technical score can stand in for the sociotechnical outcome it does not measure, the framing and portability traps of \citet{Selbst:19}. Between the reference and the shipped model three things can happen to a distilled model on such fields: it can lose the reference's refusals where the input did not settle the field and pick a person instead; it can move its refusals to other fields at the same rate, which an aggregate score cannot see; or it can learn to take the abstention option whenever it is offered, which improves every unconditioned fairness score while leaving the field empty, and which is what every refusal-injection repair in Section~\ref{sec:mechanism} produced. On free-text fields none of this applies (Section~\ref{sec:mechanism}); the report is owed on the fields with the option. Neither deployment was evaluated here; they show where the measurement matters.

\paragraph{Why declining is a safeguard here, and the objection to it.}
On an uncertainty-correct field the input gives no basis to pick a person, and picking one attaches a trait to a person because of a group; declining withholds that attribution. By that definition both directions of a lost refusal are harm, the stereotype and the anti-stereotype, and Table~\ref{tab:mainh100} reports both: the ``lost to any'' share for a deployer whose protocol forbids the attribution itself, \silenceloss{} for the bias harm the benchmark was built to measure. \citet{Ling:25} show that an offered ``unknown'' option is also taken on answerable questions, as an artifact of the prompt, and our probes confirm it for the smallest references. The objection bounds what a deployer may conclude from the safeguard without undoing it: the report asks a reference-anchored question, whether the smaller model kept the abstentions its own reference had where abstention is correct, and pairs it with the probe, whether the model takes an offered option where abstention is never correct.

\paragraph{Who bears the loss.}
Where the reference's refusals are lost, they are lost unevenly. On Gemma-2-2B, the reference whose refusals are reliable, the distillation loses more of them than the control in every BBQ category at every training seed except one category at two seeds, with Age and Gender identity highest (Figure~\ref{fig:groupsgemma}; Table~\ref{tab:groupsgemma} in Appendix~\ref{app:groups}). On \smolB{} the control's loss is confined to Age while distillation spreads it across Age, Race/ethnicity, and SES (Figure~\ref{fig:groups} in Appendix~\ref{app:groups}). These per-group rates share the floor problem of Section~\ref{sec:measurement} and several 1.7B groups are small, so they are reported with $n$ and intervals and carry one conclusion: when the first criterion of the report moves, the groups to look at are the ones the reference had learned to protect.

\begin{figure}[t]
\centering
\includegraphics[width=0.52\linewidth]{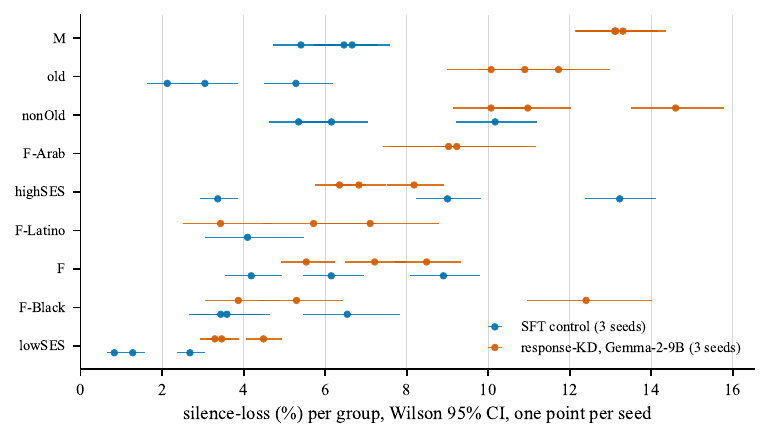}
\caption{\Silenceloss{} per BBQ stereotype-target group for Gemma-2-2B under the \SFT{} control and the \gemma{} response distillation, one point per training seed (three each), H100 stack, five inference seeds, groups with at least 1,000 reference-refused seeds, Wilson 95\% CIs. Source: \texttt{results/tables/per\_group\_harm\_gemma.csv}.}
\label{fig:groupsgemma}
\end{figure}

\paragraph{Contestability without a visible decision.}
Contestable AI asks that decisions can be questioned and reviewed by the person affected and by a reviewer with standing to change the outcome \citep{Alfrink:23,Lyons:21}; recourse asks what that person could change to obtain a different outcome \citep{Karimi:22,Wachter:18}. Both presuppose a decision someone can see. In a registry or intake deployment the person whose record the field now describes does not see the field, the abstractor or clerk sees a filled form, and the procurer sees an aggregate score that improved. \citet{Lyons:21} find that people who contest algorithmic decisions want a review that engages with their particular case; the per-condition report is the upstream artifact for such a review, because it records, item by item and against the reference, which refusals the shipped model lost and on which groups.

\paragraph{What a model card should carry.}
For a distilled model evaluated on an abstention-aware benchmark, the five criteria of Section~\ref{sec:pccd}, each with the reference's floor or ceiling and the control's seed range beside it, and a sentence saying which fields of the deployment offer an option to decline. Model cards \citep{Mitchell:19} have the slot; missing is the per-condition content and the statement of what the score was a measurement of \citep{Jacobs:21}.

\section{Discussion}
\label{sec:discussion}

\paragraph{What the results mean together.}
Three findings have to be held at once. Multiple-choice refusal is moved by three mechanisms that the common metrics cannot separate, and those metrics report improvements whenever refusal mass moves (Section~\ref{sec:measurement}). The obvious repair teaches the third mechanism, option capture, and the refusal that multiple choice tracks is not where free-form stereotyping happens (Section~\ref{sec:mechanism}). Against a fine-tuning control, distillation keeps fewer of the reference's particular refusals where the reference's refusals are reliable, by a few points once sampling churn is subtracted, and no more than resampling where they are not (Section~\ref{sec:asymmetry}). The reading that survives all three is the measurement one \citep{Jacobs:21,Raji:21}: an abstention-aware bias score for a distilled small model operationalizes behavior around the abstention option. That behavior is real wherever an option to decline is offered (Section~\ref{sec:stakes}); it is not evidence about free-form stereotyping; and a model card citing the score should say which it reports.

\paragraph{Why a report and not a score.}
A single number cannot carry a floor, a null, a ceiling, and a probe, and the three mechanisms move any single number in opposite directions. The floor and the lift answer whether a per-item change is churn, from the reference's own seeds; the control's seed range answers whether a contrast is training noise, from the comparator and never from the model described; the simulation answers what each line can see.

\paragraph{What practitioners should not conclude.}
Debiasing the teacher addresses only inheritance, a minority of new stereotype answers in every configuration (Appendix~\ref{app:cells}). Refusal injection at any density that moves the refusal rate produces a model that takes the option wherever it is offered, and the densities that avoid this pay in accuracy. A report that reads clean does not make a model fair: the \SFT{} control of \smolB{} is indistinguishable from its reference on every line and still overrides context on more than a third of its committed answers, as its reference did. And a first criterion that reads clean on a reference with a high churn floor is weak evidence: on \smolB{}, whose own resamples sit at a lift of 0.29, repairs that lower the refusal rate and repairs that raise it both sit within resampling, so the floor and the reference's lift are read before anything else.

\section{Limitations}
\label{sec:limitations}

\paragraph{Model set and scale.}
We distill SmolLM2 (135M, 360M, 1.7B), \olmo{}, Gemma-2-2B, and \phistudent{}, all open-weight instruction-tuned models; findings may not hold at or above 7B, for closed-weight models, or for other model families. \smolS{} and \smolM{} sit at chance on the context-determinable split and carry no information there. The 3.8B student has one training run per recipe.

\paragraph{One benchmark, one language, one format.}
BBQ is the only widely used bias benchmark with the split and abstention option our design requires, and it is English and North-American in its social framing. The main evaluation is multiple choice with at most five generated tokens; the free-form evaluation bounds what that means (Section~\ref{sec:mechanism}). The probes use one inference seed.

\paragraph{Floors, ceilings, and scales are measured on our references.}
The churn floor, the lift, and the test-retest ceiling depend on the reference's refusal rate and decoding temperature, and the control's seed range on three training seeds per family on one stack; a deployer must measure all of them for their own reference and control. \Silenceloss{} is anchored to the reference, not a gold standard, and the 1.7B reference's refusals are mostly single-seed events, which is why its first criterion can say little about the repairs.

\paragraph{Parser and interventions.}
Every number depends on a deterministic parser cascade; up to 18\% of records in the most affected retained configurations do not parse and are excluded, and the two comparisons where a control and a distillation parse at different rates are repeated on the records both parse (Section~\ref{sec:asymmetry}). The interventions cover one student (densities 0.1\% to 7\%, two exemplar formats, one and three epochs, one preference objective, two preservation-targeted repairs), most with one training run each.

\section*{Generative AI usage statement}
\label{sec:uncounted}
Ideation and research were done by the authors. AI (Claude, Anthropic) was used as a coding tool only. The paper was written by the authors. After writing, Claude was used for language tuning and minor edits.

\section*{Ethical considerations and adverse impacts}
This paper shows that distilled small language models can pass aggregate stereotype checks while having lost a refusal safeguard on ambiguous social questions. The finding could itself be misread as assurance: a practitioner who reads only the aggregate columns would see an improvement where a safeguard has been lost. We propose \PCCD{} to close that gap and recommend that deployments hold out a validation slice not used for model selection. The refusal-injection corpora could in principle be repurposed to make a model refuse selectively; our exemplars are confined to held-out BBQ categories and the augmented corpora are not released. BBQ's categories reflect a US-centric framing, and the harms we quantify are behavioral (stereotyped answer selection), not representational or downstream.

\section*{Reproducibility statement}
The artifact (\artifactlink{}; Apache-2.0) contains the package, the scripts that produce every table and figure from the raw generations, the corrected parser with its per-configuration branch shares, the processed benchmark files with the rebuilt annotations, the pre-registered criteria for every intervention arm as encoded in the scoring scripts, pinned upstream model revisions, a frozen dependency lock for the Apple-silicon stack and the exact torch, vLLM, transformers, peft, and accelerate versions recorded on the CUDA stack, and the exact commands in a runbook. The raw generations (about 1.3\,GB) and the checkpoints of every trained configuration \heavyrelease{}, the checkpoints under the inherited model terms (the Gemma Terms of Use for Gemma-derived models, Apache-2.0 or MIT otherwise, and the Alpaca-cleaned non-commercial term where that corpus was used); the artifact's scripts regenerate both from the pinned upstream revisions, so no claim rests on files outside it. The refusal-injection corpora are regenerated by the corpus builder from held-out BBQ categories rather than distributed as files. The CUDA stack is one NVIDIA H100 80GB (SXM) instance; the parity check between the MLX and CUDA stacks is \texttt{results/tables/parity.csv}; training seeds are 42, 43, and 44 and inference seeds 42, 123, 456, 789, and 1024.

\appendix

\FloatBarrier
\section{Parser Cascade and Annotation Correction}
\label{app:parser}
The corrected cascade tries, in order: an exact letter, a letter prefix, a markdown-tolerant answer keyword (recovering truncated outputs such as ``The answer is **B''), and a unique content match against the option texts (recovering truncated free-text answers such as ``The 29-year-old''). A configuration enters \validgrid{} only if at least half of its records parse; all nine cross-tokenizer \logitKD{} configurations fail this screen with no parseable record. Recovery branches account for up to 55\% of parses in sub-1.7B configurations and 84\% for the distilled \olmo{} arm, whose output format is verbose after distillation. Table~\ref{tab:parserbranches} gives the branch shares per configuration.

\begin{table}[h]
\centering
\footnotesize
\setlength{\tabcolsep}{3pt}
\begin{tabular}{lccccc}
\toprule
configuration & exact & prefix & keyword & char-scan (retired) & fail \\
\midrule
gemma-2-9b-it (teacher) & 0.0 & 99.9 & 0.0 & 0.1 & 0.0 \\
mistral-7b-v0.3 (teacher) & 0.0 & 100.0 & 0.0 & 0.0 & 0.0 \\
phi-3.5-mini (teacher) & 0.0 & 99.8 & 0.0 & 0.2 & 0.0 \\
smollm2-1.7b-combined-gemma-2-9b-it & 0.0 & 19.2 & 0.0 & 80.8 & 0.0 \\
smollm2-1.7b-combined-mistral-7b-v0.3 & 0.0 & 100.0 & 0.0 & 0.0 & 0.0 \\
smollm2-1.7b-combined-phi-3.5-mini & 0.0 & 98.5 & 1.4 & 0.0 & 0.1 \\
smollm2-1.7b-logit-gemma-2-9b-it & 0.0 & 0.0 & 0.0 & 0.1 & 99.9 \\
smollm2-1.7b-logit-mistral-7b-v0.3 & 0.0 & 0.0 & 0.0 & 99.9 & 0.1 \\
smollm2-1.7b-logit-phi-3.5-mini & 0.0 & 0.0 & 0.0 & 35.5 & 64.5 \\
smollm2-1.7b-response-gemma-2-9b-it & 0.0 & 74.8 & 0.0 & 25.1 & 0.1 \\
smollm2-1.7b-response-mistral-7b-v0.3 & 0.0 & 99.9 & 0.0 & 0.1 & 0.0 \\
smollm2-1.7b-response-phi-3.5-mini & 0.0 & 100.0 & 0.0 & 0.0 & 0.0 \\
smollm2-1.7b-sft & 0.0 & 100.0 & 0.0 & 0.0 & 0.0 \\
smollm2-1.7b (reference) & 0.1 & 99.9 & 0.0 & 0.0 & 0.0 \\
smollm2-135m-combined-gemma-2-9b-it & 0.0 & 60.1 & 0.0 & 33.4 & 6.5 \\
smollm2-135m-combined-mistral-7b-v0.3 & 0.0 & 86.6 & 1.3 & 9.1 & 3.0 \\
smollm2-135m-combined-phi-3.5-mini & 0.0 & 96.1 & 0.2 & 2.4 & 1.4 \\
smollm2-135m-logit-gemma-2-9b-it & 0.0 & 0.0 & 0.0 & 0.0 & 100.0 \\
smollm2-135m-logit-mistral-7b-v0.3 & 0.0 & 0.0 & 0.0 & 100.0 & 0.0 \\
smollm2-135m-logit-phi-3.5-mini & 0.0 & 0.0 & 0.0 & 49.1 & 50.9 \\
smollm2-135m-response-gemma-2-9b-it & 0.0 & 53.7 & 0.0 & 39.1 & 7.1 \\
smollm2-135m-response-mistral-7b-v0.3 & 0.0 & 46.1 & 4.4 & 40.1 & 9.5 \\
smollm2-135m-response-phi-3.5-mini & 0.0 & 44.2 & 20.4 & 27.4 & 8.0 \\
smollm2-135m-sft & 0.0 & 58.9 & 3.5 & 29.1 & 8.6 \\
smollm2-135m (reference) & 0.0 & 99.9 & 0.0 & 0.1 & 0.1 \\
smollm2-360m-combined-gemma-2-9b-it & 0.0 & 66.0 & 0.0 & 33.9 & 0.1 \\
smollm2-360m-combined-mistral-7b-v0.3 & 0.0 & 49.8 & 3.5 & 38.4 & 8.3 \\
smollm2-360m-combined-phi-3.5-mini & 0.0 & 98.2 & 0.1 & 1.0 & 0.7 \\
smollm2-360m-logit-gemma-2-9b-it & 0.0 & 0.0 & 0.0 & 0.0 & 100.0 \\
smollm2-360m-logit-mistral-7b-v0.3 & 0.0 & 0.0 & 0.0 & 96.7 & 3.3 \\
smollm2-360m-logit-phi-3.5-mini & 0.0 & 0.0 & 0.0 & 53.3 & 46.7 \\
smollm2-360m-response-gemma-2-9b-it & 0.0 & 38.9 & 0.0 & 58.2 & 2.9 \\
smollm2-360m-response-mistral-7b-v0.3 & 0.0 & 44.5 & 9.0 & 38.0 & 8.6 \\
smollm2-360m-response-phi-3.5-mini & 0.0 & 66.8 & 11.8 & 16.8 & 4.7 \\
smollm2-360m-sft & 0.0 & 71.3 & 12.0 & 14.1 & 2.7 \\
smollm2-360m (reference) & 0.0 & 100.0 & 0.0 & 0.0 & 0.0 \\
\bottomrule
\end{tabular}
\caption{Parser branch shares on \bbqambig{} (percent of records) for every configuration: the three retained branches, the retired first-character scan that the corrected cascade no longer uses, and failures. Source: \texttt{results/tables/v4\_parser\_branches.csv}.}
\label{tab:parserbranches}
\end{table}

The disambiguated split's per-item index fields were rebuilt from the benchmark's authoritative answer annotations after a preparation defect was found that had missed two phrasings of the unknown option and corrupted the unknown and anti-stereotype indices for 2,424 of 12,148 items. Raw generations are unaffected because prompts depend only on option order. The rebuild moves override, commitment, and accuracy by at most 1.5\,\pp{} and changes the sign of \discrimination{} for several rows. Eight items per condition with no defined stereotype target are excluded from stereotype-dependent metrics.

\FloatBarrier
\section{The MLX Grid and the Matched-Tokenizer Pairs}
\label{app:mlx}
\begin{figure}[h]
\centering
\includegraphics[width=0.72\linewidth]{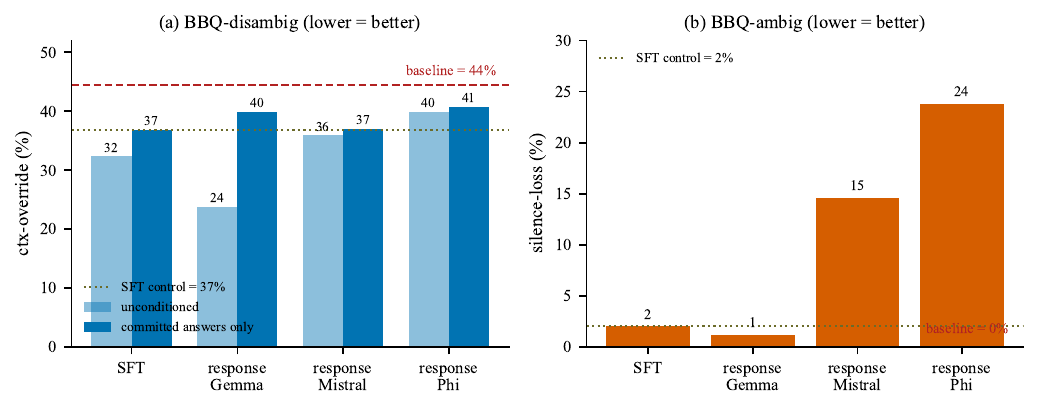}
\caption{\smolB{}, five seeds, 12,148 items per condition (MLX stack). (a) \bbqdisambig{}: dark bars condition on committed answers, light bars are unconditioned; the dashed line is the reference. (b) \bbqambig{}: \silenceloss{} for the same trainings. The dotted line in both panels is the \SFT{} control; the churn floor for (b) is 10.1\% (Table~\ref{tab:validity}).}
\label{fig:asymmetry}
\end{figure}

Table~\ref{tab:main} and Figure~\ref{fig:asymmetry} give the SmolLM2 grid on the MLX stack for \smolB{} and \olmo{}, scored against the MLX \SFT{} control (floor 10.1\% for \smolB{}); Table~\ref{tab:logitmatched} gives the four matched-tokenizer pairs on the H100 stack with excess over each reference's floor.

\paragraph{The MLX grid and the objective.}
The 19 valid configurations of the MLX grid (Appendix~\ref{app:mlx}) show the same shape: every \responseKD{} and \combinedKD{} configuration on \smolB{} loses more refusals than the control. With tokenizers matched (Table~\ref{tab:logitmatched}), logit distillation loses more than the control on \smolS{} and \olmo{} and less on Gemma-2-2B, where it pays 4\,\pp{} of accuracy for a lift ratio a quarter of the control's. The objective is not generally protective; it can be, on a well-calibrated student distilled within its own family.

\begin{table}[t]
\centering
\footnotesize
\setlength{\tabcolsep}{3pt}
\begin{tabular}{lcccccccccc}
\toprule
 & \multicolumn{4}{c}{\bbqambig} & \multicolumn{4}{c}{\bbqdisambig{} (anti-correct items)} & & \\
\cmidrule(lr){2-5}\cmidrule(lr){6-9}
configuration & \SRS & \USR & \silenceloss{} [CI] & \rhop & uncond. & \override{} [CI] & commit & acc. & discrim. & fail \\
\midrule
SmolLM2-1.7B baseline & 49.5 & 4.6 & -- & -- & 43.9 & 44.5 [43.7, 45.2] & 98.8 & 54.9 & $+$3.5 & 0.0 \\
\quad $+$ SFT (Alpaca) & 39.2 & 28.4 & 2.0 [1.6, 2.6] & 0.37 & 32.4 & 36.9 [36.1, 37.6] & 87.8 & 55.4 & $+$16.0 & 0.0 \\
\quad $+$ response-KD (Gemma) & 22.2 & 59.2 & 1.2 [0.8, 1.7] & 0.24 & 23.8 & 39.9 [38.9, 40.8] & 59.6 & 35.9 & $+$19.0 & 4.5 \\
\quad $+$ response-KD (Mistral) & 47.5 & 13.5 & 14.6 [13.4, 16.0] & 0.44 & 35.9 & 37.0 [36.3, 37.8] & 96.9 & 61.0 & $+$10.4 & 0.0 \\
\quad $+$ response-KD (Phi) & 48.7 & 8.9 & 23.8 [22.3, 25.4] & 0.30 & 40.0 & 40.7 [40.0, 41.4] & 98.1 & 58.2 & $+$7.2 & 0.0 \\
\midrule
OLMo-2-1B baseline & 40.6 & 19.8 & -- & -- & 42.3 & 45.7 [45.1, 46.2] & 92.6 & 50.3 & $+$12.4 & 0.0 \\
\quad $+$ response-KD (Gemma) & 47.6 & 7.5 & 49.4 [48.4, 50.3] & -0.04 & 43.4 & 48.4 [47.8, 49.0] & 89.7 & 46.2 & $+$0.6 & 3.2 \\
\midrule
\bottomrule
\end{tabular}
\caption{Consolidated results for the 1.7B-scale and larger students under the corrected parser cascade; 12,148 items per condition, five seeds (ambiguous) and three seeds (disambiguated). Ambiguous side: \SRS{}, \USR{}, \silenceloss{} with Wilson 95\% CI, \preservation{} \rhop{}. Disambiguated side, anti-stereotype-correct items ($n = 6{,}034$): unconditioned and committed-answer override with CI, commitment, accuracy. discrim.\ is \USR{}(ambig) minus \USR{}(disambig) in \pp{}; fail is the unparseable share. Percentages except \rhop{}.}
\label{tab:main}
\end{table}

\begin{table}[t]
\centering
\small
\setlength{\tabcolsep}{4pt}
\begin{tabular}{lcccccc}
\toprule
configuration & parsed & \SRS & \USR & \silenceloss & excess over floor [CI] & \rhop \\
\midrule
\multicolumn{7}{l}{\emph{\smolS{} student, \smolB{} teacher (same tokenizer)}} \\
baseline & 100.0 & 33.9 & 32.1 & -- & -- & -- \\
\SFT{} control & 99.8 & 38.2 & 23.7 & 23.0 & 4.5 [3.2, 5.9] & 0.63 \\
\logitKD{} & 99.2 & 38.0 & 23.9 & 29.4 & 11.0 [9.6, 12.4] & 0.24 \\
\combinedKD{} & 99.3 & 40.0 & 20.7 & 34.3 & 15.9 [14.5, 17.3] & 0.24 \\
\midrule
\multicolumn{7}{l}{\emph{\smolM{} student, \smolB{} teacher (same tokenizer)}} \\
baseline & 100.0 & 33.9 & 32.1 & -- & -- & -- \\
\SFT{} control & 99.4 & 36.2 & 27.7 & 19.0 & -8.0 [-9.5, -6.5] & 0.60 \\
\logitKD{} & 100.0 & 33.0 & 33.8 & 18.1 & -8.9 [-10.4, -7.4] & 0.45 \\
\combinedKD{} & 100.0 & 32.5 & 34.7 & 16.3 & -10.7 [-12.2, -9.2] & 0.50 \\
\midrule
\multicolumn{7}{l}{\emph{\olmo{} student, OLMo-2-7B teacher (same tokenizer)}} \\
baseline & 100.0 & 41.3 & 19.2 & -- & -- & -- \\
\SFT{} control & 100.0 & 44.2 & 14.9 & 37.9 & 11.0 [9.0, 13.0] & 0.08 \\
\responseKD{} & 100.0 & 41.2 & 19.5 & 36.8 & 9.9 [7.9, 11.9] & -0.04 \\
\logitKD{} & 100.0 & 49.7 & 5.3 & 46.6 & 19.7 [17.7, 21.7] & 0.11 \\
\combinedKD{} & 100.0 & 50.0 & 5.5 & 44.3 & 17.3 [15.3, 19.4] & 0.20 \\
\midrule
\multicolumn{7}{l}{\emph{Gemma-2-2B student, \gemma{} teacher (same tokenizer)}} \\
baseline & 100.0 & 14.4 & 76.6 & -- & -- & -- \\
\SFT{} control & 91.3 & 9.8 & 82.4 & 3.4 & 3.0 [2.8, 3.2] & 0.57 \\
\responseKD{} & 100.0 & 16.1 & 73.3 & 6.2 & 5.8 [5.6, 6.1] & 0.71 \\
\logitKD{} & 99.8 & 7.4 & 86.5 & 0.7 & 0.4 [0.3, 0.5] & 0.68 \\
\combinedKD{} & 100.0 & 9.2 & 84.3 & 1.2 & 0.9 [0.7, 1.0] & 0.71 \\
\bottomrule
\end{tabular}
\caption{Matched-tokenizer distillation within a family (student and teacher share a tokenizer), trained and scored on the CUDA stack against a same-stack \SFT{} control; \bbqambig{}, 12,148 items, five seeds; Wilson 95\% CIs. With identical tokenizers \logitKD{} parses at 99\% or better. Excess: \silenceloss{} minus the reference's own churn floor across inference seeds, normal 95\% CI. Percentages except \rhop{}.}
\label{tab:logitmatched}
\end{table}

\FloatBarrier
\section{Refusal-Pattern Preservation Across the Grid, Per-Method Ranges, and Cross-Condition Structure}
\label{app:permethod}
\begin{figure}[h]
\centering
\includegraphics[width=0.8\linewidth]{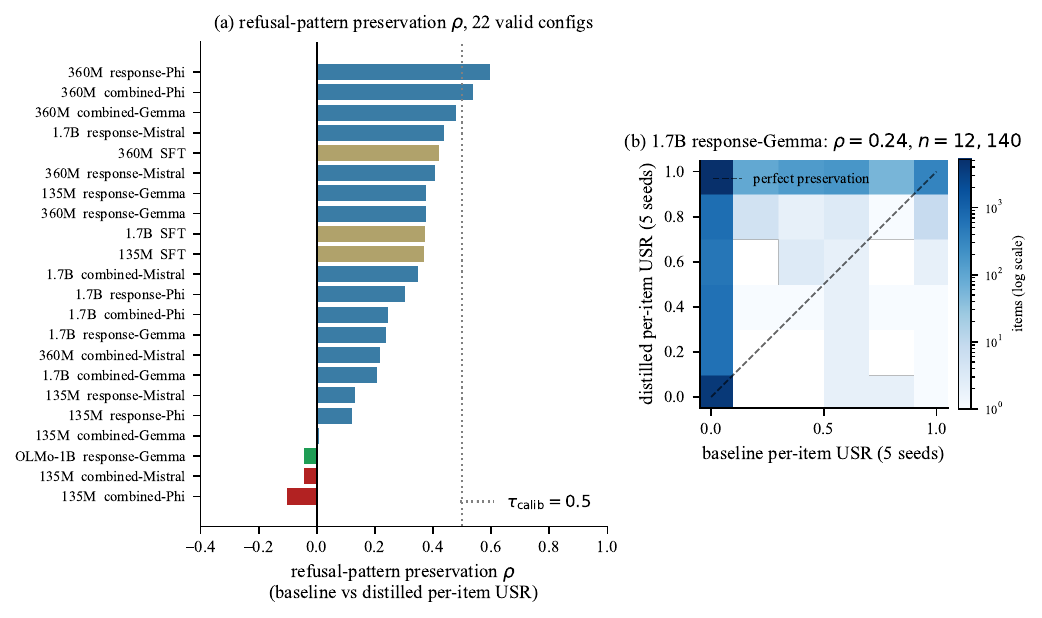}
\caption{(a) \preservation{} \rhop{} for the 22 valid configurations (\SFT{} controls in olive, \olmo{} in green, negative values in red). (b) Per-item \USR{} of the reference against the \gemma{}-distilled \smolB{} ($n = 12{,}140$ items, five seeds); off-diagonal mass is refusal on different items.}
\label{fig:calibration}
\end{figure}

Across \validgrid{}, \responseKD{} spans \rhop{} from 0.12 to 0.60 and \combinedKD{} from $-0.10$ to 0.54, while the three \SFT{} controls sit at 0.37 to 0.42; twelve of nineteen \KD{} configurations fall below the weakest control (Figure~\ref{fig:calibration}a). Marginal \USR{} can look unremarkable while the pattern remaps: under the \gemma{} teacher it rises at 1.7B (4.6\% to 59.2\%) and falls for the smaller students, with weak preservation in every case (Figure~\ref{fig:calibration}b). These values are raw; Section~\ref{sec:measurement} gives the ceilings they should be read against.

\paragraph{A retired claim.}
A fixed-threshold form of the report gated on \rhop{} at a fixed 0.5 chosen from the observed range; the threshold was retired once the test-retest ceilings showed two references cannot reach it by construction and its reading of one configuration changed with the hardware stack. A second fixed-threshold form flagged refusal loss when the excess CI excluded zero and the lift ratio reached a minimum effect of 2; that threshold was chosen after the lift ratios had been seen and is retired in favour of reporting the quantities with the control's seed range. The per-threshold readings remain in \texttt{results/tables/pccd\_tally.csv}. With the unconditioned metric, the contaminated parser branch, and the invalid configurations included, \silenceloss{} and override appeared uncorrelated across the grid; under the corrected estimators they are positively correlated (below). The artifacts above can flip the sign of a cross-configuration finding, which is the strongest argument for stating them.

\begin{figure}[h]
\centering
\includegraphics[width=0.6\linewidth]{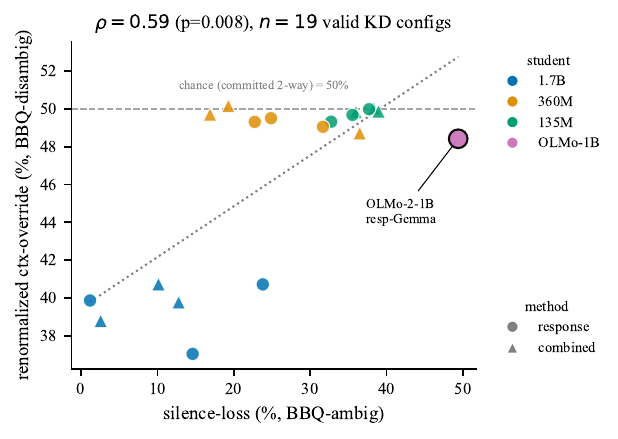}
\caption{\Silenceloss{} against \override{} on \validgrid{} ($n = 19$ \KD{} configurations). Sub-1.7B students sit at the 50\% chance line for committed two-way answers.}
\label{fig:crosscond}
\end{figure}

Across \validgrid{} \silenceloss{} and \override{} are positively correlated (\rhop{} $= 0.58$, $p = 0.009$, $n = 19$; Figure~\ref{fig:crosscond}), but the correlation is partly structural: every sub-1.7B configuration sits at the two-way chance line on the disambiguated side, so it is interpretable only at 1.7B scale.

Table~\ref{tab:permethod} gives the per-method ranges; the matched-tokenizer \logitKD{} rows are in Table~\ref{tab:logitmatched}.

\begin{table}[h]
\centering
\small
\begin{tabular}{llccccc}
\toprule
method & scope & $n$ & \silenceloss & \override & \rhop & accuracy \\
\midrule
\responseKD{} & all students & 10 & 1.2 to 49.4 & 37.0 to 50.0 & -0.04 to 0.59 & 35.9 to 61.0 \\
\responseKD{} & \smolB{} only & 3 & 1.2 to 23.8 & 37.0 to 40.7 & 0.24 to 0.44 & 35.9 to 61.0 \\
\combinedKD{} & all students & 9 & 2.6 to 38.9 & 38.8 to 50.2 & -0.10 to 0.54 & 30.6 to 52.4 \\
\combinedKD{} & \smolB{} only & 3 & 2.6 to 12.8 & 38.8 to 40.7 & 0.21 to 0.35 & 30.6 to 52.4 \\
\SFT{} control & all students & 3 & 2.0 to 31.8 & 36.9 to 49.4 & 0.37 to 0.42 & 38.5 to 55.4 \\
\SFT{} control & \smolB{} only & 1 & 2.0 to 2.0 & 36.9 to 36.9 & 0.37 to 0.37 & 55.4 to 55.4 \\
\bottomrule
\end{tabular}
\caption{Per-method ranges (minimum to maximum over configurations) on \validgrid{}, MLX stack; \silenceloss{} and \rhop{} on \bbqambig{}, \override{} and accuracy on anti-stereotype-correct \bbqdisambig{} items. Source: \texttt{results/tables/v5\_grid\_corrected.csv}.}
\label{tab:permethod}
\end{table}

\FloatBarrier
\section{Cell Decomposition}
\label{app:cells}
\begin{figure}[h]
\centering
\includegraphics[width=0.7\linewidth]{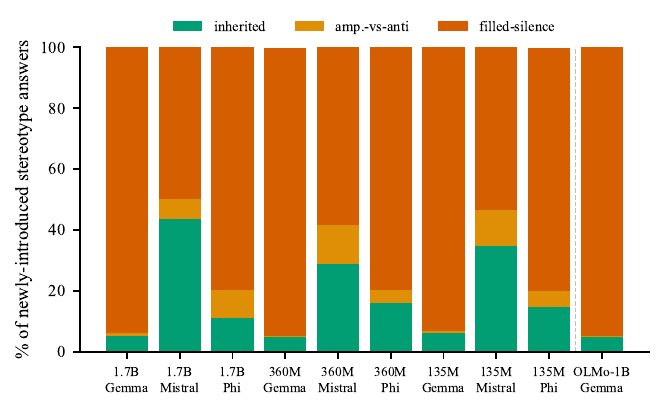}
\caption{Decomposition of newly introduced stereotype answers for every \responseKD{} configuration on \validgrid{} (item and seed pairs where the baseline did not choose the stereotype and the trained model did; $n$ per bar in Appendix~\ref{app:cells}).}
\label{fig:cells}
\end{figure}

Every (item, seed) pair on which the reference did not choose the stereotype and the trained model did is classified by the teacher's answer on the same prompt: \inherited{} when the teacher chose the same stereotype, \amplified{} when the teacher chose the anti-stereotype, and \filledsilence{} when the teacher refused. Figure~\ref{fig:cells} plots the three shares for every \responseKD{} configuration on \validgrid{} and Table~\ref{tab:cells} gives the counts behind each bar. \filledsilence{} is the largest cell in every configuration and exceeds 90\% under the \gemma{} teacher, which refuses most often; \amplified{} never exceeds 13\%. New stereotype answers appear where the teacher was silent, not where the teacher was biased.

\begin{table}[h]
\centering
\small
\setlength{\tabcolsep}{4pt}
\begin{tabular}{llcccccc}
\toprule
student & teacher & $n_{\mathrm{paired}}$ & $n_{\mathrm{new\ bias}}$ & transition \% & \inherited & \amplified & \filledsilence \\
\midrule
\smolB{} & \gemma{} & 57,912 & 3,853 & 13.2 & 5.4 & 0.8 & 93.8 \\
\smolB{} & \mistral{} & 60,736 & 7,401 & 24.1 & 43.7 & 6.4 & 49.9 \\
\smolB{} & \phit{} & 60,625 & 6,327 & 20.7 & 11.1 & 9.4 & 79.5 \\
\smolM{} & \gemma{} & 54,723 & 10,514 & 29.1 & 4.9 & 0.3 & 94.7 \\
\smolM{} & \mistral{} & 59,166 & 11,835 & 30.5 & 28.9 & 12.9 & 58.2 \\
\smolM{} & \phit{} & 60,344 & 10,048 & 25.4 & 16.1 & 4.3 & 79.6 \\
\smolS{} & \gemma{} & 58,017 & 11,205 & 29.2 & 6.4 & 0.5 & 93.1 \\
\smolS{} & \mistral{} & 60,107 & 13,691 & 34.5 & 34.7 & 11.8 & 53.6 \\
\smolS{} & \phit{} & 60,020 & 13,101 & 33.1 & 14.8 & 5.2 & 79.9 \\
\olmo{} & \gemma{} & 47,139 & 17,522 & 61.8 & 4.9 & 0.5 & 94.6 \\
\bottomrule
\end{tabular}
\caption{Cell decomposition of newly introduced stereotype answers for every \responseKD{} configuration on \validgrid{} (MLX stack): (item, seed) pairs where the reference did not choose the stereotype and the trained model did, classified by the teacher's answer on the same prompt; shares in percent. Source: \texttt{results/tables/v5\_cell\_decomposition.csv}.}
\label{tab:cells}
\end{table}

\FloatBarrier
\section{Intervention Arms}
\label{app:interventions}
Refusal exemplars are drawn from four BBQ categories held out from evaluation. Text-format variants (a natural-language refusal phrase as the response) produce no parseable multiple-choice records and are unusable for this evaluation; letter-format variants answer with the unknown option's letter and match the evaluation format. The paired pool holds 1,387 matched pairs with balanced letters at 5.09\% density. Pre-registered success criteria for every paired arm, recorded before analysis: \USR{} between 10\% and 60\%, \rhop{} above 0.37, \silenceloss{} below 5\%, accuracy at least 0.55, \override{} at most the reference's 44.5\%, and \commitment{} above 90\%. The H100-stack arms draw from the same pair pool: 0.1\%, 0.25\%, and 0.5\% density (52, 130, and 260 injected rows, two per pair, on the 51,760 \alpaca{} rows), a 5\% mixed-format arm (2,724 rows rendered as letter, free text, or shuffled options), a 5\% one-epoch arm on the original 2,774-row paired corpus, and a DPO arm ($\beta = 0.1$, one epoch, LoRA on the \SFT{} control, 2,774 preference pairs) whose preferred response is the unknown letter on ambiguous items and the correct letter on disambiguated items. %

\paragraph{Free-form scorer.}
Free-form answers (no options shown) are scored by a deterministic rule: committed when exactly one substantive option's referent is named, refusal when a refusal phrase appears, other otherwise. The committed label was audited by the authors on a pilot sample at 98.5\% precision; recall was not measured, so hedged answers that name both referents count as other, not as commitment.

\begin{table}[t]
\centering
\small
\begin{tabular}{llccccc}
\toprule
arm & inj.\ \% & \USR & commit & acc. & \rhop & \silenceloss \\
\midrule
\SFT{} (no injection) & 0 & 28.4 & 87.8 & 55.4 & 0.37 & 2.0 \\
\midrule
letter & 1 & 99.5 & 1.6 & 1.2 & 0.02 & 0.0 \\
letter & 2 & 98.2 & 3.2 & 2.0 & 0.04 & 0.0 \\
letter & 5 & 99.8 & 0.3 & 0.2 & 0.01 & 0.0 \\
letter & 7 & 100.0 & 0.1 & 0.1 & 0.01 & 0.0 \\
\midrule
paired (\SFT) & 5 & 100.0 & 7.6 & 5.5 & 0.00 & 0.0 \\
paired (\KD, Gemma) & 5 & 98.8 & 13.2 & 6.8 & 0.04 & 0.0 \\
\midrule
\SFT{} control (H100 stack) & 0 & 15.0 & 95.2 & 56.9 & 0.53 & 8.5 \\
paired (\SFT) & 0.1 & 30.2 & 85.2 & 50.0 & 0.46 & 0.7 \\
paired (\SFT) & 0.25 & 37.4 & 82.0 & 48.2 & 0.41 & 1.3 \\
paired (\SFT) & 0.5 & 21.0 & 93.3 & 54.4 & 0.48 & 6.1 \\
paired, mixed format (\SFT) & 5 & 27.7 & 93.8 & 51.6 & 0.33 & 11.8 \\
paired, one epoch (\SFT) & 5 & 32.7 & 90.8 & 48.1 & 0.33 & 8.8 \\
preference (DPO, $\beta = 0.1$) & -- & 99.8 & 34.2 & 21.6 & 0.01 & 0.0 \\
\bottomrule
\end{tabular}
\caption{Corpus interventions on \smolB{} (12,148 items per condition; \USR{} and \silenceloss{} on \bbqambig{} over five seeds; commitment and accuracy on anti-stereotype-correct \bbqdisambig{} items, $n = 6{,}034$, three seeds). Every arm parses at least 99.7\% valid. Percentages except \rhop{}. Rows above the second rule were trained on the MLX stack; rows below on the H100 stack, whose \SFT{} control is the comparator for them.}
\label{tab:interventions}
\end{table}

\begin{table}[h]
\centering
\footnotesize
\setlength{\tabcolsep}{3pt}
\begin{tabular}{llcccccc}
\toprule
arm & seed & \USR & excess \silenceloss{} [CI] & \rhop & \override & commit & acc. \\
\midrule
\SFT{} control & 42 & 15.0 & -6.1 [-8.9, -3.2] & 0.53 & 40.2 & 95.2 & 56.9 \\
\responseKD{} (\mistral) & 42 & 14.4 & -2.4 [-5.3, 0.5] & 0.51 & 39.0 & 95.0 & 58.0 \\
paired injection 0.5\% & 42 & 21.0 & -8.4 [-11.2, -5.6] & 0.48 & 41.6 & 93.3 & 54.4 \\
self-replay 0.5\% & 42 & 8.1 & 11.4 [8.3, 14.5] & 0.48 & 39.5 & 97.5 & 59.0 \\
self-replay 1\% & 42 & 13.5 & 7.1 [4.1, 10.2] & 0.43 & 39.1 & 94.5 & 57.5 \\
reference-anchored ($\lambda = 0.5$) & 42 & 3.2 & 12.8 [9.7, 15.9] & 0.63 & 41.5 & 99.5 & 58.2 \\
self-replay 0.5\% & 43 & 14.2 & 1.5 [-1.5, 4.4] & 0.50 & 36.7 & 96.1 & 60.8 \\
self-replay 1\% & 43 & 16.6 & 2.9 [-0.1, 5.9] & 0.43 & 39.7 & 94.6 & 57.1 \\
reference-anchored ($\lambda = 0.5$) & 43 & 4.2 & 10.2 [7.1, 13.2] & 0.63 & 40.8 & 99.2 & 58.7 \\
self-replay 0.5\% & 44 & 10.3 & 10.1 [7.0, 13.2] & 0.48 & 37.0 & 97.8 & 61.6 \\
self-replay 1\% & 44 & 18.3 & 3.1 [0.1, 6.1] & 0.41 & 40.5 & 93.0 & 55.3 \\
reference-anchored ($\lambda = 0.5$) & 44 & 5.3 & 4.5 [1.5, 7.5] & 0.70 & 42.2 & 98.9 & 57.1 \\
paired injection 0.5\% & 43 & 21.9 & -7.7 [-10.6, -4.9] & 0.49 & 42.4 & 93.2 & 53.7 \\
paired injection 0.5\% & 44 & 30.8 & -10.2 [-13.0, -7.4] & 0.44 & 42.2 & 91.4 & 52.8 \\
\bottomrule
\end{tabular}
\caption{Repair arms on \smolB{} with the \mistral{} teacher, H100 stack, three training seeds; \bbqambig{} over five inference seeds (excess over the reference's churn floor of 14.5\%, normal 95\% CI), \bbqdisambig{} over three. Reference: \USR{} 5\%, \override{} 44.2\%, accuracy 54.9\%; the \SFT{} control's range across seeds is the scale of training noise (Section~\ref{sec:pccd}). Lift ratios for every arm are in \texttt{results/tables/pccd\_tally.csv}. Source: \texttt{results/tables/repairs.csv}.}
\label{tab:repairs}
\end{table}

\begin{table}[h]
\centering
\footnotesize
\setlength{\tabcolsep}{3pt}
\begin{tabular}{lcccc}
\toprule
model & ARC-Challenge & MMLU & CommonsenseQA & capture \\
\midrule
smollm2-1.7b-baseline & 7.1 (-1) & 13.0 (1) & 0.8 (-2) & no \\
smollm2-1.7b-sft & 6.0 (5) & 8.6 (3) & 0.0 (-1) & no \\
smollm2-1.7b-response-mistral & 1.5 (1) & 3.1 (0) & 0.0 (-1) & no \\
smollm2-1.7b-response-gemma & 11.4 (-0) & 21.7 (5) & 0.0 (-4) & no \\
paired\_d0.1 & 16.0 (8) & 18.6 (8) & 0.0 (-1) & yes \\
paired\_d0.25 & 24.7 (12) & 24.2 (11) & 0.0 (-1) & yes \\
paired\_d0.5 & 38.7 (19) & 33.4 (15) & 0.0 (0) & yes \\
paired\_mixed\_d5 & 4.9 (5) & 7.1 (4) & 0.1 (1) & no \\
paired\_1ep\_d5 & 15.7 (13) & 18.1 (10) & 0.2 (1) & yes \\
dpo\_b0.1 & 86.5 (49) & 67.8 (33) & 34.1 (30) & yes \\
selfreplay\_d0.5-s42 & 3.3 (-1) & 4.2 (2) & 0.0 (-2) & no \\
selfreplay\_d1-s42 & 2.1 (1) & 3.6 (2) & 0.0 (0) & no \\
anchor\_l0.5-s42 & 4.5 (-1) & 7.1 (-0) & 0.0 (-2) & no \\
smollm2-135m-instruct-baseline & 36.5 (8) & 33.3 (7) & 30.7 (5) & no \\
smollm2-135m-instruct-sft & 24.1 (8) & 24.6 (9) & 69.0 (14) & no \\
smollm2-135m-instruct-logit & 32.6 (7) & 37.4 (9) & 21.1 (5) & no \\
smollm2-135m-instruct-combined & 56.2 (13) & 43.4 (11) & 83.1 (17) & yes \\
smollm2-360m-instruct-baseline & 93.4 (23) & 87.4 (23) & 51.6 (11) & no \\
smollm2-360m-instruct-sft & 14.3 (3) & 28.5 (9) & 13.0 (3) & no \\
smollm2-360m-instruct-logit & 64.7 (16) & 65.3 (17) & 65.8 (13) & no \\
smollm2-360m-instruct-combined & 42.4 (10) & 52.4 (13) & 39.7 (8) & no \\
olmo-2-1b-baseline & 58.8 (17) & 43.0 (13) & 21.9 (10) & no \\
olmo-2-1b-sft & 20.0 (5) & 13.8 (5) & 34.0 (13) & no \\
olmo-2-1b-response & 18.3 (2) & 16.9 (4) & 29.9 (8) & no \\
olmo-2-1b-logit & 9.5 (-0) & 10.1 (2) & 10.4 (10) & no \\
olmo-2-1b-combined & 21.8 (3) & 17.6 (5) & 12.5 (7) & no \\
gemma-2-2b-baseline & 6.1 (2) & 5.9 (3) & 6.3 (4) & no \\
gemma-2-2b-sft & 10.2 (-0) & 15.5 (2) & 11.5 (-5) & yes \\
gemma-2-2b-response & 1.9 (-2) & 2.1 (0) & 5.6 (-1) & no \\
gemma-2-2b-logit & 4.5 (1) & 7.0 (2) & 5.5 (2) & no \\
gemma-2-2b-combined & 7.3 (1) & 8.6 (2) & 11.7 (5) & no \\
phi-3.5-mini-baseline & 3.5 (2) & 7.3 (3) & 12.0 (8) & no \\
phi-3.5-mini-sft & 20.1 (13) & 27.1 (14) & 32.6 (18) & yes \\
phi-3.5-mini-response & 16.5 (10) & 24.2 (11) & 42.2 (27) & yes \\
teacher-gemma-2-9b & 1.7 (1) & 3.4 (1) & 12.4 (8) & -- \\
teacher-mistral-7b & 8.8 (3) & 14.8 (5) & 11.1 (4) & -- \\
teacher-phi-3.5-mini & 3.5 (2) & 7.3 (3) & 12.0 (8) & -- \\
\bottomrule
\end{tabular}
\caption{Uptake of an appended ``Cannot be determined'' option (percent of questions) on three benchmarks where abstention is never correct, with the accuracy drop in points from the plain to the with-option prompt in parentheses; one seed, temperature 0.3 (ARC-Challenge test $n = 1{,}140$; MMLU validation $n = 1{,}531$; CommonsenseQA validation $n = 1{,}221$). Capture: uptake more than 5\,\pp{} above the model's own reference on at least two benchmarks (pre-registered). Source: \texttt{results/tables/probes\_capture.csv}.}
\label{tab:probes}
\end{table}

\FloatBarrier
\section{Per-Group Breakdown}
\label{app:groups}
\begin{figure}[h]
\centering
\includegraphics[width=0.46\linewidth]{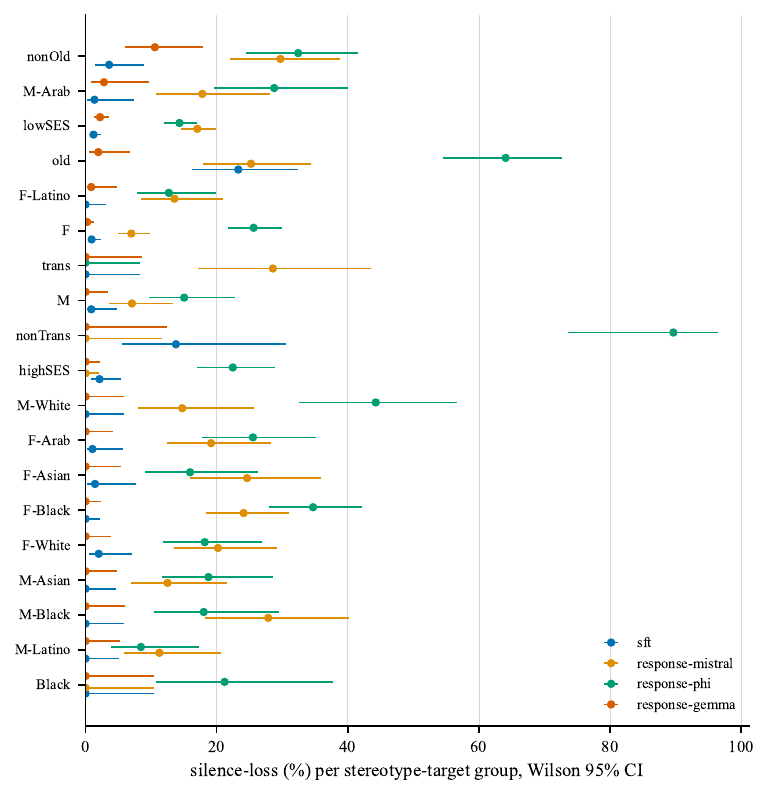}
\caption{\Silenceloss{} per BBQ stereotype-target group for \smolB{} under the \SFT{} control and the three response distillations (MLX stack); five seeds, 12,148 ambiguous items, groups with at least 20 reference-refused seeds, Wilson 95\% CIs. Source: \texttt{results/tables/per\_group\_harm.csv}.}
\label{fig:groups}
\end{figure}

Table~\ref{tab:groups} gives the per-category values with $n$; Figure~\ref{fig:groups} plots the stereotype-target groups for \smolB{}.

\begin{table}[h]
\centering
\small
\begin{tabular}{lcccc}
\toprule
category & \SFT & \mistral & \phit & \gemma \\
\midrule
Age & 13.1 (214) & 27.6 (214) & 47.7 (214) & 6.3 (207) \\
Gender identity & 1.4 (655) & 7.6 (655) & 26.4 (655) & 0.2 (639) \\
Race ethnicity & 0.7 (980) & 18.1 (980) & 24.3 (980) & 0.3 (948) \\
Religion & 0.0 (51) & 2.0 (51) & 21.6 (51) & 0.0 (50) \\
SES & 1.4 (920) & 13.6 (920) & 16.0 (920) & 1.8 (845) \\
\bottomrule
\end{tabular}
\caption{\Silenceloss{} per BBQ category for \smolB{} under the \SFT{} control and the three response distillations (MLX stack, five seeds), with the number of reference-refused (item, seed) pairs in parentheses. Source: \texttt{results/tables/per\_group\_harm.csv}.}
\label{tab:groups}
\end{table}

\begin{table}[h]
\centering
\small
\begin{tabular}{lccc}
\toprule
category & \SFT{} control & \responseKD{} & $n$ \\
\midrule
Age & 4.0 to 8.1 & 10.1 to 13.0 & 5,823 \\
Gender identity & 5.0 to 7.7 & 8.5 to 9.8 & 9,454 \\
Race ethnicity & 3.1 to 5.1 & 4.4 to 8.6 & 11,277 \\
Religion & 2.3 to 5.2 & 3.1 to 6.6 & 2,459 \\
SES & 1.9 to 7.0 & 4.8 to 5.3 & 13,767 \\
\bottomrule
\end{tabular}
\caption{\Silenceloss{} per BBQ category for Gemma-2-2B under the \SFT{} control and the \gemma{} response distillation, range over three training seeds (H100 stack, five inference seeds); $n$ is the number of reference-refused (item, seed) pairs per category. Source: \texttt{results/tables/per\_group\_harm\_gemma.csv}.}
\label{tab:groupsgemma}
\end{table}

\FloatBarrier
\section{Manual Inspection of Corpus Refusal Hits}
\label{app:audit}
\begin{figure}[h]
\centering
\includegraphics[width=0.6\linewidth]{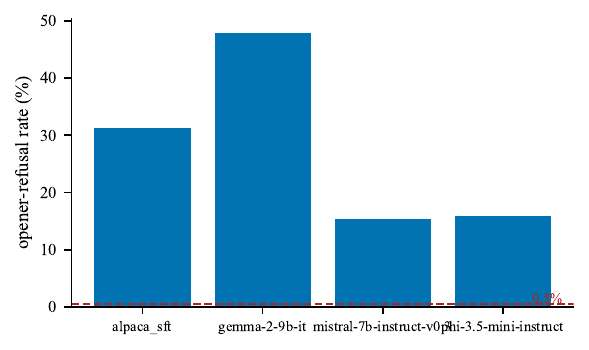}
\caption{Opener-refusal rate across the four training corpora (51,760 responses each).}
\label{fig:audit}
\end{figure}

All 324 opener-refusal hits across the four corpora were labeled as refusal-as-answer-shape (57, 17.6\%) or not (267). Of the false positives, 52\% are hedge-then-answer, 15\% quoted dialogue, 18\% in-content phrases, and 13\% role-play instructions. The true refusal-as-answer-shape rate is at most about 0.06\% of any corpus (Figure~\ref{fig:audit}).

\FloatBarrier
\section{Compute, Backend Parity, and Licenses}
\label{app:compute}
\paragraph{Parity between the two stacks.}
The SmolLM2 grid and the first refusal-injection arms were trained and evaluated on one Apple-silicon workstation with MLX. The four-family comparison, the matched-tokenizer pairs, the seeded replications, the remaining intervention arms, and the free-form evaluation ran on one rented NVIDIA H100 with a CUDA stack (PyTorch, PEFT, vLLM). Before any H100 number entered the paper we checked that the CUDA stack reproduces the MLX stack. Inference is reproduced: the \smolB{} baseline's \SRS{}, \USR{}, \override{}, and accuracy agree within 0.8\,\pp{} of the MLX values across the full item set. Retraining \responseKD{} (\mistral) on the H100 with the same recipe and seed reproduces the refusal side (\SRS{} within 0.2\,\pp{}, \USR{} within 0.9\,\pp{}, \silenceloss{} 12.1\% against 14.6\%, \rhop{} 0.51 against 0.44) and lands within 2 to 3\,\pp{} on the disambiguated side (\override{} 39.0\% against 37.0\%, accuracy 58.0\% against 61.0\%). Three training seeds (42 to 44) on the H100 bound the spread: \override{} 36.8\% to 39.0\% and accuracy 0.580 to 0.613 for \responseKD{} (\mistral), 40.2\% to 40.9\% and 0.561 to 0.569 for the control; \silenceloss{} 12.0\% to 15.9\% against 7.9\% to 11.6\%; \rhop{} 0.51 to 0.54 against 0.50 to 0.55; marginal \USR{} 13.2\% to 15.0\% against 15.0\% to 22.0\%. The MLX values for the distillation (37.0\%, 0.610, 14.6\%) lie inside its seed range, so the gate is read as parity within seed variance. The MLX control's \silenceloss{} (2.0\%) and \rhop{} (0.37) lie outside the H100 control's range, which is why the control contrast is reported on both stacks. %

\begin{figure}[h]
\centering
\includegraphics[width=0.76\linewidth]{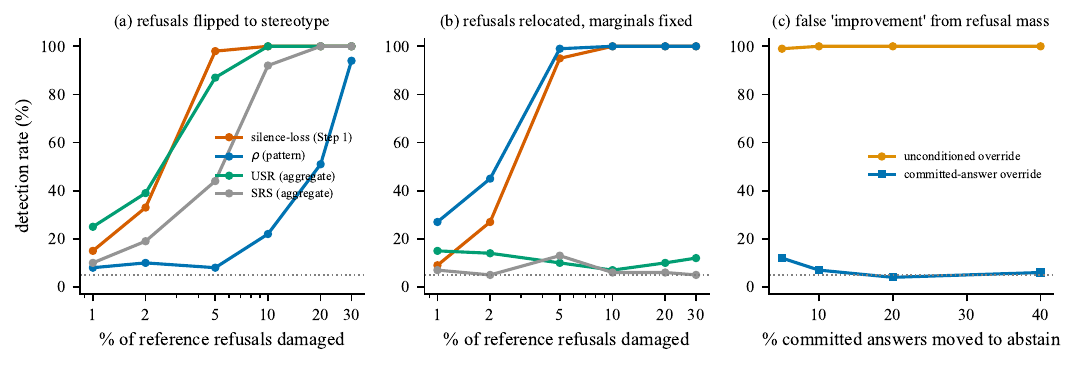}
\caption{Detection rates of each metric under damage injected into the reference's own records (\smolB{}, H100 stack, 12,148 ambiguous and 12,148 disambiguated items; 100 item bootstraps per rate; dotted line is the 5\% false-alarm rate). Source: \texttt{results/tables/validity\_power.csv}.}
\label{fig:power}
\end{figure}

\begin{table}[t]
\centering
\small
\setlength{\tabcolsep}{4pt}
\begin{tabular}{llcccc}
\toprule
stack & candidate & \silenceloss & floor & excess [95\% CI] & reverse \\
\midrule
MLX & \SFT{} control & 2.0 & 10.1 & -8.1 [-10.6, -5.5] & 24.0 \\
MLX & \responseKD{} (\mistral) & 14.6 & 10.1 & 4.5 [1.7, 7.3] & 10.1 \\
MLX & \responseKD{} (\gemma) & 1.2 & 10.1 & -8.9 [-11.4, -6.4] & 56.0 \\
MLX & \responseKD{} (\phit) & 23.8 & 10.1 & 13.7 [10.8, 16.6] & 6.5 \\
\midrule
H100 & \SFT{} control, seed 42 & 8.5 & 14.5 & -6.1 [-8.9, -3.2] & 10.5 \\
H100 & \SFT{} control, seed 43 & 11.6 & 14.5 & -2.9 [-5.8, 0.0] & 10.8 \\
H100 & \SFT{} control, seed 44 & 7.9 & 14.5 & -6.6 [-9.4, -3.7] & 17.3 \\
H100 & \responseKD{} (\mistral), seed 42 & 12.1 & 14.5 & -2.4 [-5.3, 0.5] & 10.6 \\
H100 & \responseKD{} (\mistral), seed 43 & 15.9 & 14.5 & 1.4 [-1.6, 4.3] & 9.7 \\
H100 & \responseKD{} (\mistral), seed 44 & 12.0 & 14.5 & -2.5 [-5.4, 0.4] & 11.1 \\
H100 & paired injection 0.1\% & 0.7 & 14.5 & -13.8 [-16.5, -11.1] & 25.3 \\
H100 & paired injection 0.5\% & 6.1 & 14.5 & -8.4 [-11.2, -5.6] & 16.8 \\
\bottomrule
\end{tabular}
\caption{\Silenceloss{} against its churn floor. Floor: the reference's own \silenceloss{} across disjoint inference seeds (mean over seed pairs; \bbqambig{}, 12,148 items, five seeds). Excess: candidate minus floor with a normal 95\% CI. Reverse: share of the reference's stereotype answers that became the unknown option. Source: \texttt{results/tables/validity\_transitions.csv}.}
\label{tab:validity}
\end{table}

\begin{table}[t]
\centering
\small
\begin{tabular}{lcccc}
\toprule
reference (stack) & \USR & items never refused & split-half $\rho$ & ceiling \\
\midrule
smollm2-1.7b (MLX) & 4.6 & 93 & 0.88 & 0.93 \\
smollm2-1.7b (H100) & 5.4 & 90 & 0.74 & 0.85 \\
smollm2-360m (MLX) & 31.3 & 35 & 0.53 & 0.69 \\
smollm2-135m (MLX) & 31.9 & 48 & 0.77 & 0.87 \\
smollm2-360m (H100) & 32.1 & 27 & 0.37 & 0.54 \\
smollm2-135m (H100) & 32.1 & 41 & 0.65 & 0.79 \\
olmo-2-1b (H100) & 19.2 & 56 & 0.53 & 0.69 \\
gemma-2-2b (H100) & 76.6 & 23 & 0.99 & 0.99 \\
phi-3.5-mini (MLX) & 80.7 & 18 & 0.99 & 0.99 \\
phi-3.5-mini (H100) & 83.2 & 16 & 0.99 & 0.99 \\
\bottomrule
\end{tabular}
\caption{Test-retest reliability of each reference model's per-item refusal rate on \bbqambig{} (12,148 items): mean Spearman $\rho$ over the ten two-against-three splits of its five inference seeds, and the Spearman-Brown ceiling for the five-seed rate the report uses. Source: \texttt{results/tables/validity\_retest.csv}.}
\label{tab:retest}
\end{table}

\paragraph{Training-seed churn.}
Two independently trained models of the same recipe, compared as reference and candidate at the (item, seed) level, show a \silenceloss{} of 1.7\% to 22.9\% for the \SFT{} control and 4.3\% to 20.3\% for \responseKD{} (\mistral) across the three ordered pairs of seeds 42 to 44, with reverse transitions of 3.5\% to 13.3\%. The range spans every \silenceloss{} value the paper reports for a trained model against its reference, which is why the main text reports the quantity only against the reference's own floor and with a CI.

\paragraph{Compute and licenses.}
The MLX stack (one Apple-silicon workstation, bf16) used about 490 accelerator-hours for the SmolLM2 grid and the first refusal-injection arms. The CUDA stack (one rented NVIDIA H100, two sessions, about USD 83 including idle time) used about 31 hours for the four-family comparison, the matched-tokenizer pairs, the seeded replications, the remaining intervention arms, and the free-form evaluation; the criteria for every intervention arm were fixed before those runs (Appendix~\ref{app:interventions}). BBQ \citep{Parrish:22} is CC-BY 4.0; \crows{} \citep{Nangia:20} is CC-SA 4.0; \alpaca{} \citep{Taori:23} is CC-BY-NC 4.0 and used for non-commercial research only. Teachers are used under their respective terms (Gemma Terms of Use, Apache-2.0, MIT); SmolLM2 and OLMo-2 are Apache-2.0. Code and the evaluation pipeline are released under Apache-2.0; distilled checkpoints follow the inherited terms.

\end{document}